\documentclass[11pt, dvipsnames]{article}

\usepackage[preprint]{sty/acl/acl}

\usepackage{times}
\usepackage{latexsym}

\usepackage[T1]{fontenc}

\usepackage[utf8]{inputenc}

\usepackage{microtype}

\usepackage{inconsolata}

\usepackage[utf8]{inputenc} 
\usepackage[T1]{fontenc}    
\usepackage{hyperref}       
\usepackage{url}            
\usepackage{booktabs}       
\usepackage{amsfonts}       
\usepackage{nicefrac}       
\usepackage{microtype}      
\usepackage{xcolor}         
\usepackage{array}

\usepackage{caption}
\usepackage{multirow}
\usepackage{makecell}
\usepackage{algorithm}
\usepackage[noend]{algpseudocode}
\usepackage{seqsplit}
\usepackage{minibox}
\usepackage{booktabs}       
\usepackage{amsfonts}       
\usepackage{nicefrac}       
\usepackage{amssymb}
\usepackage{enumitem} 
\usepackage{wrapfig}
\usepackage{tabularx}
\usepackage{tikz}
\usetikzlibrary{shapes,backgrounds,arrows,fit,calc,patterns}
\usepackage{graphicx}
\usepackage{subcaption}
\usepackage{xcolor}

\newcommand{\ie}{\emph{i.e.,}~}
\newcommand{\eg}{\emph{e.g.,}~}

\usepackage{pifont}

\usepackage{amsmath} 
\usepackage{bm} 
\usepackage{bbm} 
\usepackage{amsthm}
\usepackage{mathtools}

\newcommand{\ep}{\varepsilon}
\renewcommand{\epsilon}{\ep}

\makeatletter
\@ifundefined{proposition}{%
    
}{}
\@ifundefined{definition}{%
}{}
\@ifundefined{lemma}{%
    
}{}
\@ifundefined{corollary}{%
    \newtheorem{corollary}{Corollary}
}{}
\@ifundefined{theorem}{%
    
}{}
\@ifundefined{corollary}{%
    
}{}
\@ifundefined{assumption}{%
    
}{}
\@ifundefined{problem}{%
    
}{}
\@ifundefined{problem*}{%
    \newtheorem*{problem*}{Problem}
}{}
\@ifundefined{claim}{%
    
}{}
\@ifundefined{example}{%
    
}{}
\@ifundefined{implication}{%
    
}{}
\@ifundefined{remark*}{%
    \newtheorem*{remark*}{Remark}
}{}
\@ifundefined{observation}{%
    \newtheorem{observation}{Observation}
}{}
\makeatother

\newtheorem{innercustomgeneric}{\customgenericname}
\providecommand{\customgenericname}{}
\newcommand{\newcustomtheorem}[2]{%
  \newenvironment{#1}[1]
  {%
   \renewcommand\customgenericname{#2}%
   \renewcommand\theinnercustomgeneric{##1}%
   \innercustomgeneric
  }
  {\endinnercustomgeneric}
}
\newcustomtheorem{customclaim}{Claim}

\usepackage[capitalise]{cleveref}
\usepackage{longtable}

\title{The Interplay of Harness Design and Post-Training in LLM Agents}

\author{%
  \textbf{Kyungmin Kim}$^{1,*}$,
  \textbf{Youngbin Choi}$^{1,*}$,
  \textbf{Seoyeon Lee}$^1$,
  \textbf{Suhyeon Jun}$^2$,\\
  \textbf{Dongwoo Kim}$^{1,2,\dagger}$,
  \textbf{Sangdon Park}$^{1,2,\dagger}$
  \\
 \textsuperscript{1}Graduate School of Artificial Intelligence, POSTECH,
\\
 \textsuperscript{2}Department of Computer Science and Engineering, POSTECH, 
\\
 \small{\texttt{\{kkm959595, choi.youngbin, seoyeon26, suhyeonjun, dongwoo.kim, sangdon\}@postech.ac.kr}}
}

\begin{document}

\maketitle
\begingroup
\renewcommand\thefootnote{}\footnotetext{{$^{*}$}Equal contribution.}
\renewcommand\thefootnote{}\footnotetext{$^\dagger$Co-advising.}


\newcommand{\Reg}{\textbf{Reg}}

\begin{abstract}
Tool-integrated LLM agents are often wrapped within a \emph{harness}: the scaffolding that determines which tools are exposed, how they are described, and what auxiliary information accompanies each per-step observation.
While agents are routinely post-trained, this scaffolding is typically treated as a \emph{fixed} engineering detail, with design effort limited to the training-free regime.
Moreover, existing post-training algorithms assume a static environment, even though tool environments and tasks often shift upon deployment.
To address this gap, we extend \texttt{ALFWorld} (i) to treat the harness as a controllable design dimension and (ii) to support evaluation under task and tool environment shifts. Building on this, we systematically analyze how the harness design influences post-training in both in-distribution and out-of-distribution (OOD) settings.
We empirically show that \emph{harness-aware post-training} not only improves in-distribution performance but also enables agents to robustly adapt to OOD settings. 
Under a harness with minimal design effort, post-training suffers a drastic performance drop under stronger tool environment shifts, further highlighting the importance of harness-aware post-training under such shifts.
\end{abstract}

\section{Introduction}
\label{sec:intro}
LLM agents are designed to solve complex problems that often require multi-step reasoning~\citep{yao2023react, shinn2023reflexion, wang2024voyager, yang2024swe, wu2025agentic}.
Among them, \textit{tool-integrated agentic systems} extend LLM capabilities by proactively interacting with environments through predefined tools~\citep{schick2023toolformer, qin2024toolllm, patil2024gorilla, liu2024apigen}.
Accordingly, performance depends not only on producing correct final outputs but also on appropriately invoking tools, which has motivated increasing efforts to post-train pretrained LLMs for effective tool use~\citep{feng2025retool}.

A factor that is often overlooked in this line of work is the \textit{harness}: the code that wraps the LLM, determining what is stored, retrieved, and presented to it, and what is done with its outputs~\citep{wang2025openhands, anthropic2026harness, openai2026harness}.
While the harness is, in principle, an engineering detail, in practice it is a major determinant of overall performance: even with the same task and model, two harnesses that differ only in how they present auxiliary information can produce drastically different success rates~\citep{yang2024swe,badertdinov2026swe}.
{\color{black}For this reason, harness design has attracted increasing attention, ranging from manually curated prompts~\citep{hong2024metagpt,antoniades2025swe, lin2026agentic} to recent attempts at automating the design itself~\citep{lee2026meta,shang2025agentsquare}.}
All of these efforts, however, are limited to a training-free regime, treating the harness as a way to elicit better performance, typically from a closed-source model.
How harness design interacts with post-training, in contrast, remains largely unexamined.

\begin{figure*}[t]
     \centering
     \includegraphics[width=\textwidth]{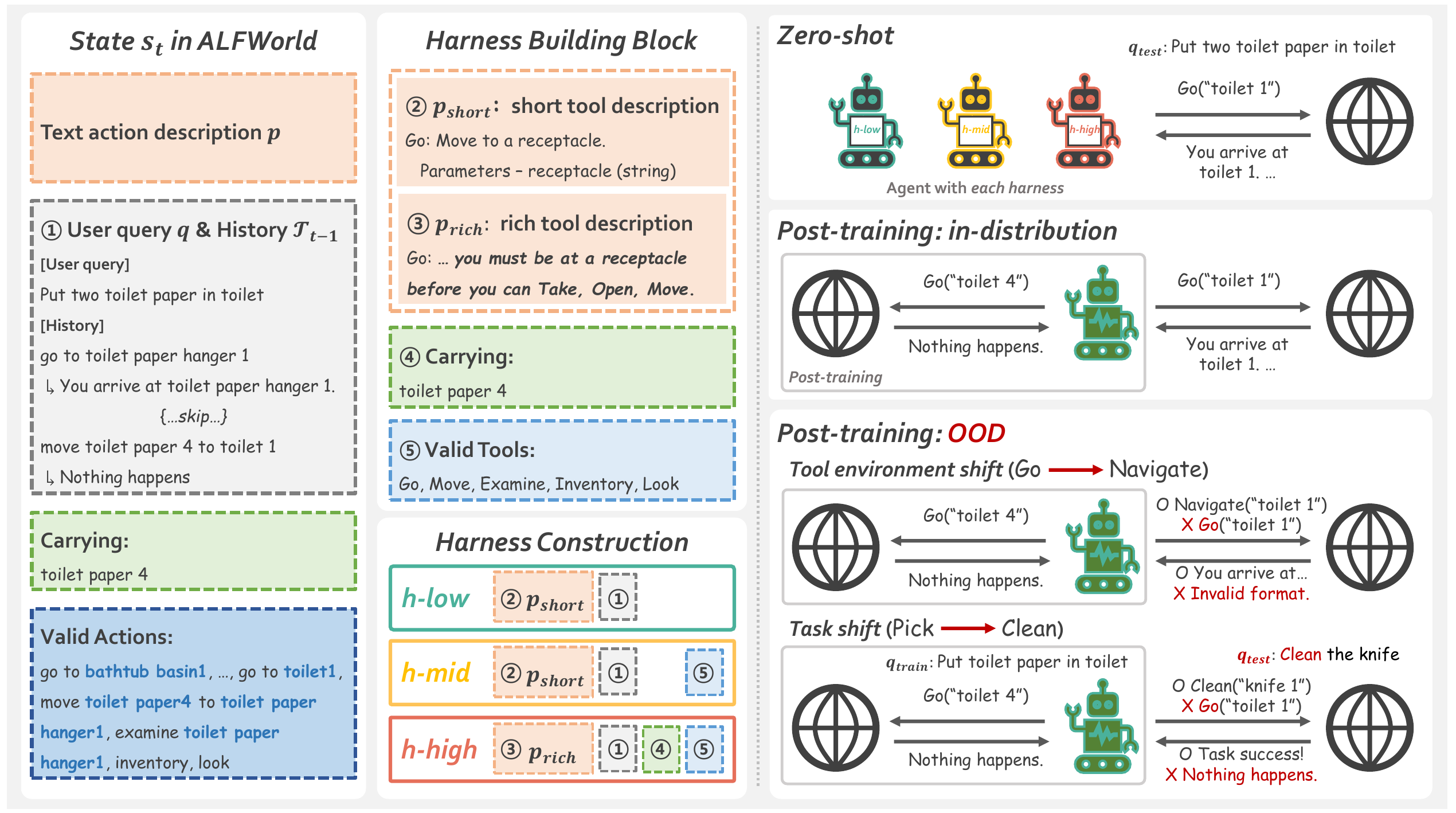}
     \caption{Overview of experimental design, built on our extension of \texttt{ALFWorld}~\citep{shridhar2021alfworld} tailored to tool-integrated agentic tasks. 
     (Left) State of the original text-based \texttt{ALFWorld} environment. (Middle) Harness building blocks for tool-integrated agentic tasks{\color{black}:} different combinations of these blocks yield harnesses with varying levels of informativeness. 
     (Right) Agents under each harness are evaluated in zero-shot and post-training regimes, where the latter spans in-distribution and OOD settings.}
     \label{fig:different_feedback_scenarios}
\end{figure*}

In parallel, most existing post-training approaches for tool-integrated agents implicitly assume a static deployment scenario, in which both the task distribution and the tool environment, defined as the set of tools and their invocation protocols, remain fixed~\citep{jin2025search, mai2025agent, xue2025simpletir, qian2025toolrl, feng2026groupingroup, jiang-etal-2025-s3, cheng2025agent}.
In practice, this assumption is often violated: real-world deployments expose agents to shifts in both the task distribution and the tool environment.
A first source of shift arises from changes in the task distribution, a standard out-of-distribution (OOD) problem at the input level~\citep{koh2021wilds, yuan2023revisiting, yang2023out}. 
For example, an agent originally designed for workload scheduling may later be repurposed for itinerary planning, shifting objectives, constraints, and tool usage patterns, which we refer to as \textit{task shift}.
A second, more subtle source stems from changes in the tool environment itself: tools may be updated, modifying invocation protocols while preserving functionality, which we refer to as \textit{tool environment shift}.
This introduces a distinct OOD setting in which the task distribution remains fixed but the action interface itself changes, and it remains comparatively underexplored~\citep{chen2025toolevo, li2026proevolve}.

In this paper, we ask the following question: \textit{how does harness design influence post-training LLM agents in both in-distribution and OOD settings?}
To this end, we extend $\texttt{ALFWorld}$~\citep{shridhar2021alfworld}, a text-based planning environment, into a benchmark tailored to tool-integrated agentic tasks.
Specifically, we (i) treat the harness as a controllable design dimension and (ii) modularize the benchmark to support evaluation both in-distribution and under tool environment shift and task shift settings (\cref{fig:different_feedback_scenarios}).
While existing post-training algorithms attain high success rates on \texttt{ALFWorld} \citep{feng2026groupingroup}, we observe that this performance largely stems from \emph{harness-aware post-training}, where a specific harness is fixed a priori (\cref{fig:iid}).
In particular, the original benchmark provides the agent with the complete set of feasible tool calls at every step, which constitutes a highly informative harness.
Such a harness, however, comes at a cost: it can be constructed only when the transition dynamics of the \texttt{ALFWorld} environment are fully specified.
Yet this cost is rarely made explicit in prior work.
To address this gap, we provide three harnesses with varying levels of informativeness and empirically examine \emph{whether the additional cost of harness design is indispensable for post-training}.

{\color{black}
We empirically show that harness-aware post-training not only improves in-distribution performance but also enables agents to robustly adapt to OOD settings. 
Notably, under the stronger tool environment shift where only the invocation protocol changes, models post-trained on a harness with low design effort suffer from a drastic performance drop, despite their decent in-distribution performance, which is even comparable to that of the agents post-trained with an informative harness.
In addition, we observe that applying an informative harness only after post-training {\color{black}under a low-effort harness} yields performance inferior to that obtained when the harness is in place during post-training.
Unlike the closed-source setting, where post-hoc harness engineering is the standard practice \cite{lee2026meta}, this result provides an actionable insight: a high-quality harness should be designed in advance and treated as an explicit part of the post-training recipe rather than as an inference-time add-on.}

\section{Preliminaries}
\label{sec:prelim}
We outline the formulation of multi-step tool-integrated agentic tasks and representative post-training methods, along with the notation used throughout the paper.

\subsection{Multi-step Tool-integrated Agentic Tasks}
{\color{black} Let $\mathcal{V}$ be a token vocabulary, $\mathcal{S} \coloneqq \mathcal{V}^\ast$ be a state space, and $\mathcal{A} \coloneqq \mathcal{V}^\ast$ be an action space.}
Given a system prompt $p \in \mathcal{V}^\ast$ and a user query $q \in \mathcal{V}^\ast$, an LLM agent $\pi : \mathcal{S} \rightarrow \mathcal{A}$ generates a sequence of action tokens $a_t \coloneqq \pi(s_t) \in {\color{black}\mathcal{A}}$ at each time step $t \in \{ 1, \dots, T \}$. 
Here, $s_t \in \mathcal{S}$ is the state at step $t$ provided to the agent, and $T$ is the total number of tool calling steps.
Specifically, $s_t \coloneqq ( p,  q, \mathcal{T}_{t-1} )$, where $\mathcal{T}_{t-1}$ is a history of tool calling steps up to the $(t-1)$-th step, and $\mathcal{T}_0$ is an empty sequence.

{\color{black} The tool environment $\texttt{TE} : \mathcal{V}^* \to \mathcal{V}^*$ then plays two roles: in the \emph{tool calling phase}, it returns the tool calling result $e_t := \texttt{TE}((s_t, a_t))$; in the \emph{state reconstruction phase}, it aggregates $e_t$ with $a_t$ and $\mathcal{T}_{t-1}$ into a history $\mathcal{T}_t := \texttt{TE}((\mathcal{T}_{t-1}, a_t, e_t))$, provided to the agent as the next state $s_{t+1} = (p, q, \mathcal{T}_t)$.}
As such, the $t$-th tool calling step can be represented as:
\begin{equation*}
    \scalebox{0.87}{$\begin{aligned}
        &\underbrace{s_{t} = ( p, q, \mathcal{T}_{t-1} )
        \xrightarrow[\scriptscriptstyle+a_t=\pi(s_t)]{}
        ( p, q, \mathcal{T}_{t-1}, a_t )
        \xrightarrow[\scriptscriptstyle+e_t=\texttt{TE}((s_t, a_t))]{}
        }_{\text{Tool Calling Phase}} \\
        &\underbrace{\rightarrow( p, q, \mathcal{T}_{t-1}, a_t, e_t )
        \xrightarrow[\substack{\scriptscriptstyle+\mathcal{T}_t=\texttt{TE}(({\color{black}\mathcal{T}_{t-1}}, a_t, e_t))}]{\scriptscriptstyle-(\mathcal{T}_{t-1}, a_t, e_t)}
        s_{t+1} = ( p, q, \mathcal{T}_{t} ).}_{\text{State Reconstruction Phase}}
    \end{aligned}$}
\end{equation*}
\vspace{-1em}
{\color{black}\paragraph{Remark.}
While ``harness'' generally refers to any wrapper around the agent \cite{lin2026agentic}, we restrict it to wrappers acting through $p$ and \texttt{TE}. 
These are the core components of the agentic workflow that shape what the agent observes in our environment.}
As detailed in \S\ref{sec:benchmark}, harness, tool schema, and task type jointly affect the transition dynamics {\color{black} of the agentic workflow}.
In particular, tool environment shift and task shift pose challenges for training an agent that remains robust to deployment-time workflows differing from those encountered during training. 

\subsection{Training Tool-integrated Agentic Systems}
We model a multi-step tool calling process as a Markov decision process (MDP) at a sequence level.
The goal is to find a policy $\pi_\theta: \mathcal{S} \rightarrow \mathcal{A}$, parameterized by $\theta \in \Theta$, that maximizes $\mathcal{J}(\theta) = \mathbbm{E} [ R( \tau ) ]$, where $\tau \coloneqq \{ ( s_1, a_1, s_2 ), \dots, ( s_T, a_T, s_{T+1} ) \}$ is an episode-level trajectory and $R: ( \mathcal{S} \times \mathcal{A} \times \mathcal{S} )^T \rightarrow \mathbbm{R}$ is an \emph{episode-level} reward function.
A common choice is $R(\tau) \coloneqq \sum_{t=1}^T \gamma^{t-1} r_t$, where $\gamma \in [0, 1]$ is a discount factor and $r_t \coloneqq r ( s_t, a_t, s_{t+1} )$ is a \emph{step-level} reward.

Group-relative policy optimization (GRPO) \citep{shao2024deepseekmath} is a widely adopted algorithm for post-training LLMs due to its computational efficiency in estimating $\nabla_\theta \mathcal{J}(\theta)$.
However, as GRPO uniformly assigns the episode-level advantage across all steps, it suffers from a credit assignment problem in long-horizon tasks \citep{dong2025agentic, fang2026proximity}.
Unlike GRPO, group-in-group policy optimization (GiGPO) \citep{feng2026groupingroup} differentiates the advantage assigned to each step within the same episode by introducing a step-level advantage on top of the episode-level advantage.
{\color{black} Yet both algorithms assume a static agentic workflow.}
A formal description of both algorithms is provided in Appendix~\ref{app.exp_details}.


{\color{black}\paragraph{Remark.}
Let $\mathcal{H}$ be a harness space.
Given a harness $h \in \mathcal{H}$ specified by $( p, \texttt{TE} )$, an agent $\pi_\theta$ is post-trained using trajectories whose per-step states and transitions are induced by the same harness $h$.
Concretely, the agent is post-trained on states constructed by that harness.
\textit{Harness-aware post-training} is then defined as post-training an LLM agent using an ``informative'' harness that encodes rich prior knowledge about the environment or the task itself, often requiring costly design effort in advance.
Existing works that adopt the aforementioned algorithms in \texttt{ALFWorld} fix a highly informative harness in advance, rather than treating it as a design variable.
As such, they lack a controlled analysis of how harness-aware post-training affects performance and OOD robustness of LLM agents.}

\section{Experiment Setup}
\label{sec:benchmark}

\subsection{\texttt{ALFWorld} for Tool-based Agentic Tasks}
\texttt{ALFWorld}~\citep{shridhar2021alfworld} is a text-based planning environment with $3,827$ task instances across six categories of common household activities: Pick \& Place (Pick), Examine in Light (Look), Clean \& Place (Clean), Heat \& Place (Heat), Cool \& Place (Cool), and Pick Two \& Place (Pick 2).
Every task requires multi-step decision making, since it consists of a sequence of sub-goals.
This structure makes \texttt{ALFWorld} a natural testbed for evaluating the planning capability of LLM agents.

We extend \texttt{ALFWorld} by reformulating its original tasks as tool-integrated reasoning problems.
Specifically, text actions in \texttt{ALFWorld} can naturally be interpreted as tool calls: verbs correspond to tools and entities serve as arguments.
For instance, a text action \texttt{``Go to drawer\,1''} in the original \texttt{ALFWorld} environment is mapped to a tool call \texttt{``Go(receptacle=`drawer\,1')''}.

In subsequent sections, we describe how our extended benchmark enables a controlled study of harness design in post-training LLM agents by modularizing three components: harness (\S\ref{sec:harness}), tool schema (\S\ref{sec:tool_schema}), and task type (\S\ref{sec:task_type}).



\begin{table}[t]
\centering

\renewcommand{\arraystretch}{1.15}
\setlength{\tabcolsep}{4pt}
\resizebox{\linewidth}{!}{
\begin{tabular}{c c c c}
\toprule
\textbf{Harness Ver.} & \makecell{\textbf{Tool Description} \\ \textbf{in $p$}}
& \makecell{\textbf{\texttt{Valid tools:}} \\ \textbf{(in $\mathcal{T}_t$)}}
& \makecell{\textbf{\texttt{Carrying:}} \\ \textbf{(in $\mathcal{T}_t$)}} \\
\midrule
\texttt{h-low} & Short & --          & --          \\
\texttt{h-mid}  & Short & \checkmark  & --          \\
\texttt{h-high} & Rich  & \checkmark  & \checkmark  \\
\bottomrule
\end{tabular}
}
\caption{Each harness version, characterized by how it shapes the system prompt $p$ and the per-step history $\mathcal{T}_t$ produced by $\texttt{TE}$. \texttt{Valid tools:} and \texttt{Carrying:} are auxiliary lines appended to $\mathcal{T}_t$, listing the tools admissible at the current state and the object the agent is currently carrying, respectively.}

\label{tab:harness_versions}
\end{table}

\subsection{Harness}
\label{sec:harness}

In tool-integrated agentic tasks, the harness determines how much guidance the agent receives about using the available tools.
To this end, we instantiate three harnesses with varying degrees of informativeness, each building on the previous one: \texttt{h-low}, \texttt{h-mid}, and \texttt{h-high}.

The base version, \texttt{h-low}, is designed with low design effort: every tool carries only a one-line description in $p$ (e.g., \texttt{``Move to a receptacle''} for \texttt{Go}), 
and each per-step history $\mathcal{T}_t$ contains nothing beyond the raw observation.
\texttt{h-mid} additionally augments every per-step history with the set of tools admissible at the current state, leaving the one-line descriptions in $p$ unchanged.
Finally, \texttt{h-high} further expands each tool description in $p$ into a richer form, covering its preconditions, interactions with other tools, and role in completing tasks. 
It also appends the object the agent is currently carrying to each per-step history, which it would otherwise have to retrieve through a separate \texttt{Inventory} call.
\cref{tab:harness_versions} summarizes the three versions, and the corresponding tool descriptions and per-step histories are deferred to \cref{app:harness_examples}.




Informative harnesses like \texttt{h-mid} and \texttt{h-high}, however, do not come for free.
They must either be hand-designed by experts with sufficient prior knowledge of the environment, 
or be discovered through costly exploratory interaction with it.
This cost is rarely made explicit. 
Most prior studies on \texttt{ALFWorld} in fact operate under a specific harness that goes beyond even \texttt{h-high}, providing the agent with the full set of feasible tool calls at every step, yet this is rarely flagged as an assumption.
We instead treat the harness as a controllable variable, with \texttt{h-low}, \texttt{h-mid}, and \texttt{h-high} corresponding to three distinct levels of informativeness.
We acknowledge that \texttt{h-mid} and \texttt{h-high} expose {\color{black} environment-derived} information that would otherwise require costly exploration to obtain.
{\color{black} 
Indeed, the \texttt{``Valid tools:''} field is an oracle-style admissibility signal computed from the \texttt{ALFWorld} simulator.
The \texttt{``Carrying:''} field is likewise state-dependent, though it could in principle be maintained by tracking previous interactions.
In addition, the rich tool descriptions in $p$ provide state-independent guidelines, which would nonetheless require expert knowledge or a sufficient amount of exploration to obtain.}
\emph{We view \texttt{h-mid} and \texttt{h-high} as representing harnesses obtained through different amounts of such exploration, treating informativeness itself as a design choice that shapes agent performance} (\textbf{RQ1} in \S\ref{sec:analysis}).



\subsection{Tool Schema}
\label{sec:tool_schema}

\begin{table}[t]
\centering
\small
\renewcommand{\arraystretch}{1.15}
\setlength{\tabcolsep}{4pt}
\resizebox{\linewidth}{!}{
\begin{tabular}{c l}
\toprule
\multicolumn{1}{c}{\textbf{Tool Schema}} & \multicolumn{1}{c}{\textbf{Example}} \\
\midrule
\texttt{v1.0} & \texttt{Go(receptacle=``drawer 1'')} \\
\texttt{v1.1} & \texttt{NavigateTo(destination=``drawer 1'')} \\
\multirow{2}{*}{\texttt{v2.0}} & \texttt{ReceptacleControl(action=``navigate\_to'',} \\
                                & \quad \texttt{target=``drawer 1'')} \\
\bottomrule
\end{tabular}
}
\caption{Tool schema versions and the valid form of tool calls for the same operation ``move to drawer 1'' under each schema. \texttt{v1.1} applies paraphrasing to \texttt{v1.0}, while \texttt{v2.0} additionally groups tools by structural and functional similarity.}
\label{tab:tool_schema_versions}
\end{table}

The tool schema determines the surface form of admissible tool calls.
Taking \texttt{v1.0} as the base schema, we provide two environment shift scenarios of varying degrees, where the stronger shift (\texttt{v2.0}) extends the milder one (\texttt{v1.1}) through additional updates to the tool invocation protocols.
Specifically, \texttt{v1.0} is the base schema itself, consisting of $13$ tools defined in \texttt{ALFWorld} with intuitive verb-style names such as \texttt{Go}, \texttt{Take}, \texttt{Heat}, \texttt{Slice}, and \texttt{Look}.
The second, \texttt{v1.1}, applies paraphrasing: a semantics-preserving renaming of every tool name and parameter key from \texttt{v1.0} into more descriptive forms (e.g., \texttt{Go} $\rightarrow$ \texttt{NavigateTo}, with its parameter \texttt{receptacle} $\rightarrow$ \texttt{destination}), leaving the cardinality of the tool set unchanged.
The third, $\texttt{v2.0}$, additionally applies grouping by structural and functional similarity: building on \texttt{v1.1}, tools that share structural or functional roles are consolidated into a smaller, higher-level set, reducing the cardinality from 13 to 5.
Each consolidated tool exposes its sub-operations through a discrete \texttt{action} parameter.
For example, \texttt{NavigateTo}, \texttt{OpenContainer}, and \texttt{CloseContainer} are merged into a single \texttt{ReceptacleControl} tool, whose \texttt{action} parameter takes one of three values: \texttt{navigate\_to}, \texttt{open\_container}, or \texttt{close\_container}.
\cref{tab:tool_schema_versions} provides an illustrative example of the three versions, with full details in \cref{app:tool_schema_details}.

Across the three versions, both $p$ and $\texttt{TE}$ are updated accordingly, with details deferred to Appendix~\ref{app:tool_schema_details}. 
Tool names, parameter keys, and descriptions in $p$ are revised to match the active schema, while the overall structure of $p$ is preserved. 
$\texttt{TE}$'s validation logic is adjusted accordingly. 
As an illustration, the same operation ``move to drawer 1'' issued at step $t$ takes a distinct valid form under each schema (\cref{tab:tool_schema_versions}). 
A tool call has a valid format only under the schema in which it is defined. 
Otherwise, even though the call expresses the same operation, \texttt{TE} returns $e_t = $ \texttt{``Invalid tool format''}.
We aim to analyze \emph{whether harness-aware post-training enables agents to robustly adapt to such shifts in tool invocation protocols} (\textbf{RQ3} in \S\ref{sec:analysis}).
\subsection{Task Type}
\label{sec:task_type}

\begin{table}[t]
\centering
\renewcommand{\arraystretch}{1.15}
\setlength{\tabcolsep}{2.5pt}
\resizebox{\linewidth}{!}{
\begin{tabular}{c c c l}
\toprule
\textbf{Group} & \textbf{Task} & \textbf{\# Sub-goals} & \multicolumn{1}{c}{\textbf{Example}} \\
\midrule
\multirow{2}{*}{\texttt{t-easy}}
    & Pick         & 4 & \texttt{``Put a plate on the coffee table''} \\
    & Look      & 3 & \texttt{``Examine a book under the lamp''} \\
\midrule
\multirow{3}{*}{\texttt{t-med}}
    & Clean        & 5 & \texttt{``Clean the knife and put in the drawer''} \\
    & Heat         & 5 & \texttt{``Heat a mug and put on the coffee table''} \\
    & Cool         & 5 & \texttt{``Put a cool bottle on the countertop''} \\
\midrule
\texttt{t-hard}
    & Pick 2     & 8 & \texttt{``Put two pencils in the drawer''} \\
\bottomrule
\end{tabular}
}
\caption{Six task categories of \texttt{ALFWorld} grouped into three difficulty levels (\texttt{t-easy}, \texttt{t-med}, \texttt{t-hard}) by the minimum number of sub-goals required for completion.}
\label{tab:task_groups}
\end{table}


The task type determines the distribution from which the user query $q$ is drawn. 
We group the six task categories of \texttt{ALFWorld} by the minimum number of sub-goals required for completion into three difficulty levels: $\texttt{t-easy}$, $\texttt{t-med}$, and $\texttt{t-hard}$ (\cref{tab:task_groups}).
For example, task instances in the Pick category (e.g., \texttt{``Put a plate on the coffee table''}) are categorized as \texttt{t-easy}, requiring a sequence of four sub-goals: find an object (\texttt{``plate''}), pick it up, find the correct location (\texttt{``coffee table''}) to place it, and put it down there.
This grouping serves as the basis for the task shift regime: training on one difficulty level and evaluating on the others constitutes a controlled shift in the input distribution, with the tool schema held fixed.

While task instances are categorized into three task groups, some sub-goals are shared across instances from different groups.
For example, every task instance shares the same initial sub-goal: to find an object of the desired type.
Because the corresponding tool calls are reused across instances (possibly with different arguments), an ideal post-training method should learn to generate such tool calls in a way that generalizes to task instances unseen during training, rather than overfitting to the training distribution.
While these household activities are simple from a human perspective, results in \S\ref{sec:analysis} show that they present non-trivial challenges to LLM agents in both in-distribution and OOD scenarios.
We further ask \emph{whether the prior knowledge encoded in an informative harness helps an agent generalize across task groups} (\textbf{RQ3} in \S\ref{sec:analysis}).

\section{Analysis}
\label{sec:analysis}
In this section, we present a controlled analysis of how harness design influences the post-training of LLM agents, both in-distribution and under OOD settings. Specifically, we investigate the following research questions:
\begin{itemize}
    \item \textbf{RQ1:} Does the influence of harness informativeness, observed at zero-shot (Obs.~\ref{obs:zero-shot}), extend to harness-aware post-training (Obs.~\ref{obs:iid})?
    
    \item \textbf{RQ2:} Can a harness be applied only after post-training, or should it be in place during training (Obs.~\ref{obs:post-hoc})?

    \item \textbf{RQ3:} Does harness-aware post-training enable robustness to tool environment shift (Obs.~\ref{obs:tes}) and task shift (Obs.~\ref{obs:ts})?
\end{itemize}

\begin{figure}[t]
    \centering
    \includegraphics[width=0.95\linewidth]{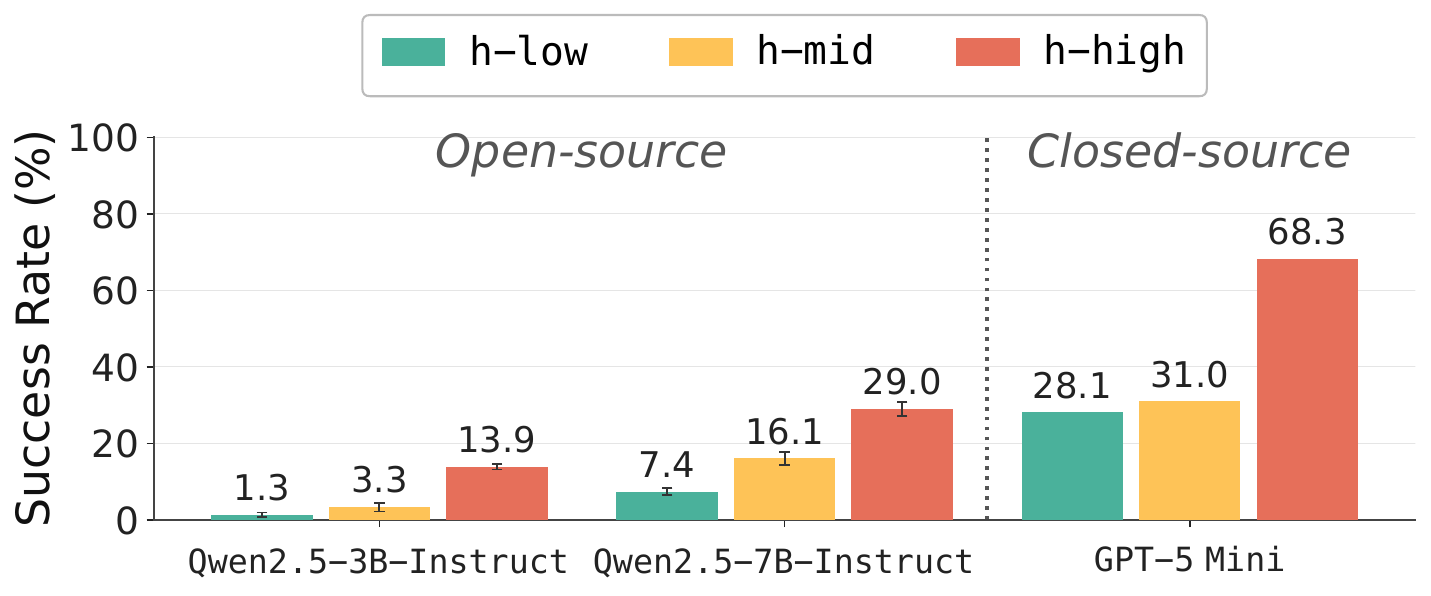}
    \caption{
        Zero-shot success rates without post-training:
        Pretrained agents are evaluated on $\mathcal{D}^{\text{te}}_{\text{all}}$ under each harness with tool schema fixed to \texttt{v1.0}.
        Zero-shot performance improves with harness informativeness, more pronounced for high-capacity models.
    }
    \label{fig:zeroshot}
\end{figure}

\subsection{Evaluation Protocol}
\label{sec:eval_protocol}
Among $3,827$ task instances in \texttt{ALFWorld}, $3,553$ form the training set $\mathcal{D}^{\text{tr}}_{\text{all}}$ and the remaining $274$ form the test set $\mathcal{D}^{\text{te}}_{\text{all}}$.
The test set corresponds to the standard seen/unseen split of \texttt{ALFWorld} ($140$ seen and $134$ unseen instances).
Both sets are partitioned by task difficulty:
$\mathcal{D}^{\text{tr}}_{\text{all}}
    \coloneqq
    \mathcal{D}^{\text{tr}}_{\text{easy}}
    \cup
    \mathcal{D}^{\text{tr}}_{\text{med}}
    \cup
    \mathcal{D}^{\text{tr}}_{\text{hard}}$
and
$\mathcal{D}^{\text{te}}_{\text{all}}
    \coloneqq
    \mathcal{D}^{\text{te}}_{\text{easy}}
    \cup
    \mathcal{D}^{\text{te}}_{\text{med}}
    \cup
    \mathcal{D}^{\text{te}}_{\text{hard}}$.
Here, $\mathcal{D}^{\text{tr}}_{\text{easy}}$, $\mathcal{D}^{\text{tr}}_{\text{med}}$, $\mathcal{D}^{\text{tr}}_{\text{hard}}$ refer to the training instances within each difficulty-based task group (\texttt{t-easy}, \texttt{t-med}, \texttt{t-hard}) defined in \S\ref{sec:task_type}.
The test splits are defined analogously.
We report the \emph{success rate}, defined as the fraction of test task instances in which the agent completes all required sub-goals within the maximum number of tool calling steps.

For \emph{zero-shot analysis}, we evaluate pretrained LLM agents directly under each of the three harnesses $\{ \texttt{h-low}, \texttt{h-mid}, \texttt{h-high} \}$ with tool schema fixed to \texttt{v1.0} (Obs.~\ref{obs:zero-shot}).
Specifically, we evaluate \texttt{GPT-5 Mini} as a closed-source model, and instruction-tuned versions of \texttt{Qwen2.5-3B} and \texttt{Qwen2.5-7B} as open-source models.
For \texttt{GPT-5 Mini}, we set its reasoning effort to ``high'' to use the most capable inference setting.

For \emph{post-training analysis}, we post-train the same open-source models with GRPO and GiGPO as RL algorithms.
For in-distribution and tool environment shift scenarios, agents are post-trained on the full training split $\mathcal{D}^{\text{tr}}_{\text{all}}$ under each of the three harnesses, with tool schema fixed to \texttt{v1.0}.
Post-trained agents are then evaluated on $\mathcal{D}^{\text{te}}_{\text{all}}$ under the same harness, either with schema \texttt{v1.0} (in-distribution, Obs.~\ref{obs:iid}) or with \texttt{v1.1} and \texttt{v2.0} corresponding to tool environment shifts of varying degrees (Obs.~\ref{obs:tes}).
For the task shift scenario, agents are instead post-trained on a single difficulty-specific split ($\mathcal{D}^{\text{tr}}_{\text{easy}}$, $\mathcal{D}^{\text{tr}}_{\text{med}}$, or $\mathcal{D}^{\text{tr}}_{\text{hard}}$) under each of the three harnesses, with tool schema fixed to \texttt{v1.0}, and evaluated on the test splits of the remaining groups under the same harness and schema (Obs.~\ref{obs:ts}).
All open-source results are averaged over three random seeds, with standard deviations reported in the corresponding figures and tables.
Additional training and evaluation details are provided in \cref{app.exp_details} and \cref{app:add_experiments}.

\paragraph{Remark.}
Prior post-training studies on \texttt{ALFWorld} typically operate under a single $(\text{harness}, \text{task type})$ configuration, training only on the full training split $\mathcal{D}^{\text{tr}}_{\text{all}}$.
We instead treat harness and task type as controllable variables, running each RL algorithm across all combinations of three harness levels and four task type settings (the full training split  and three difficulty-specific splits), for both open-source models.
This main grid of experiments alone amounts to approximately $1{,}800$ H200 GPU-hours.

\subsection{Zero-shot and In-distribution Performance (RQ1)}
\label{sec:zeroshot}
\begin{figure}[t]
    \centering
    \includegraphics[width=\linewidth]{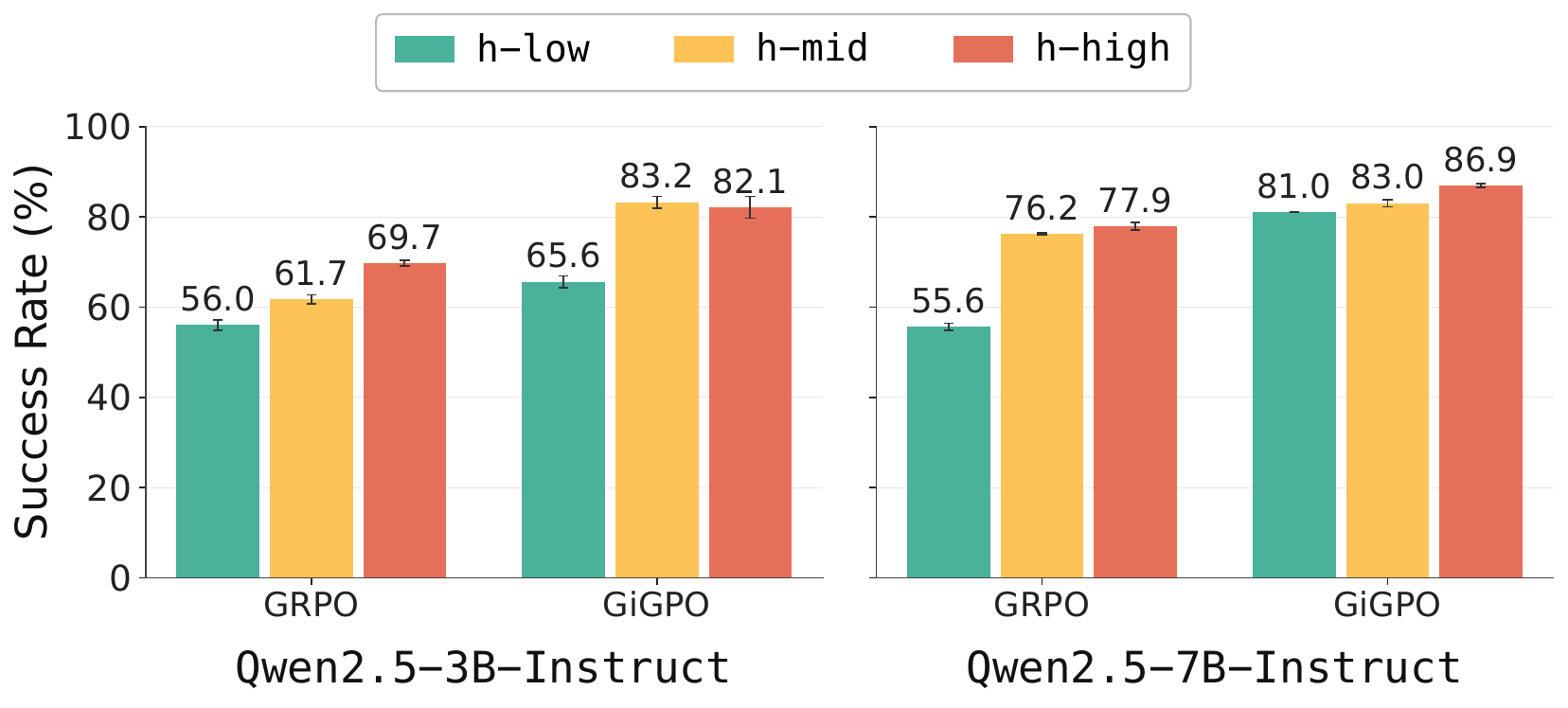}
    \caption{
        In-distribution success rates after post-training:
        {\color{black}Agents are post-trained on $\mathcal{D}_{\text{all}}^{\text{tr}}$ under each harness with tool schema fixed to \texttt{v1.0}, and evaluated on $\mathcal{D}_{\text{all}}^{\text{te}}$ under the same harness and schema.
        Post-training performance tends to improve with harness informativeness.}
    }
    \label{fig:iid}
\end{figure}


\begin{observation}[Zero-shot]
\label{obs:zero-shot}
Harness informativeness monotonically improves zero-shot performance of LLM agents, and the magnitude of the gain scales with model capacity.
\end{observation}

Across all three models, performance improves monotonically with harness informativeness (\cref{fig:zeroshot}).
In particular, \texttt{GPT-5 Mini} shows the largest gain, confirming that the magnitude of the gain scales with model capacity.
Furthermore, while open-source models fail at Pick\,2 (\texttt{t-hard}), reporting $0.0$ across most harness configurations, \texttt{GPT-5 Mini} achieves $61.0$ under \texttt{h-high} (\cref{tab:exp0_iid_zeroshot}).
Even under \texttt{h-low}, \texttt{GPT-5 Mini} achieves $17.1$ at \texttt{Pick\,2}, further illustrating that model capacity is essential to drive harness-induced gains.


\begin{observation}[In-distribution]
    \label{obs:iid}
    The monotonic harness gain observed at zero-shot (\cref{fig:zeroshot}) largely carries over after post-training under both algorithms.
\end{observation}


The monotonic harness gain observed at zero-shot {\color{black} extends} to post-training under both algorithms (\cref{fig:iid}).
For example, \texttt{Qwen2.5-3B-Instruct} post-trained with GRPO under \texttt{h-high} outperforms \texttt{Qwen2.5-7B-Instruct} post-trained with GRPO under \texttt{h-low} by $14.1$ points, indicating that the choice of harness can outweigh the effect of model capacity even after post-training.
We also observe that GiGPO consistently outperforms GRPO across all configurations, consistent with its finer credit assignment in long-horizon tasks.


\begin{figure}[t]
    \centering
    \includegraphics[width=\linewidth]{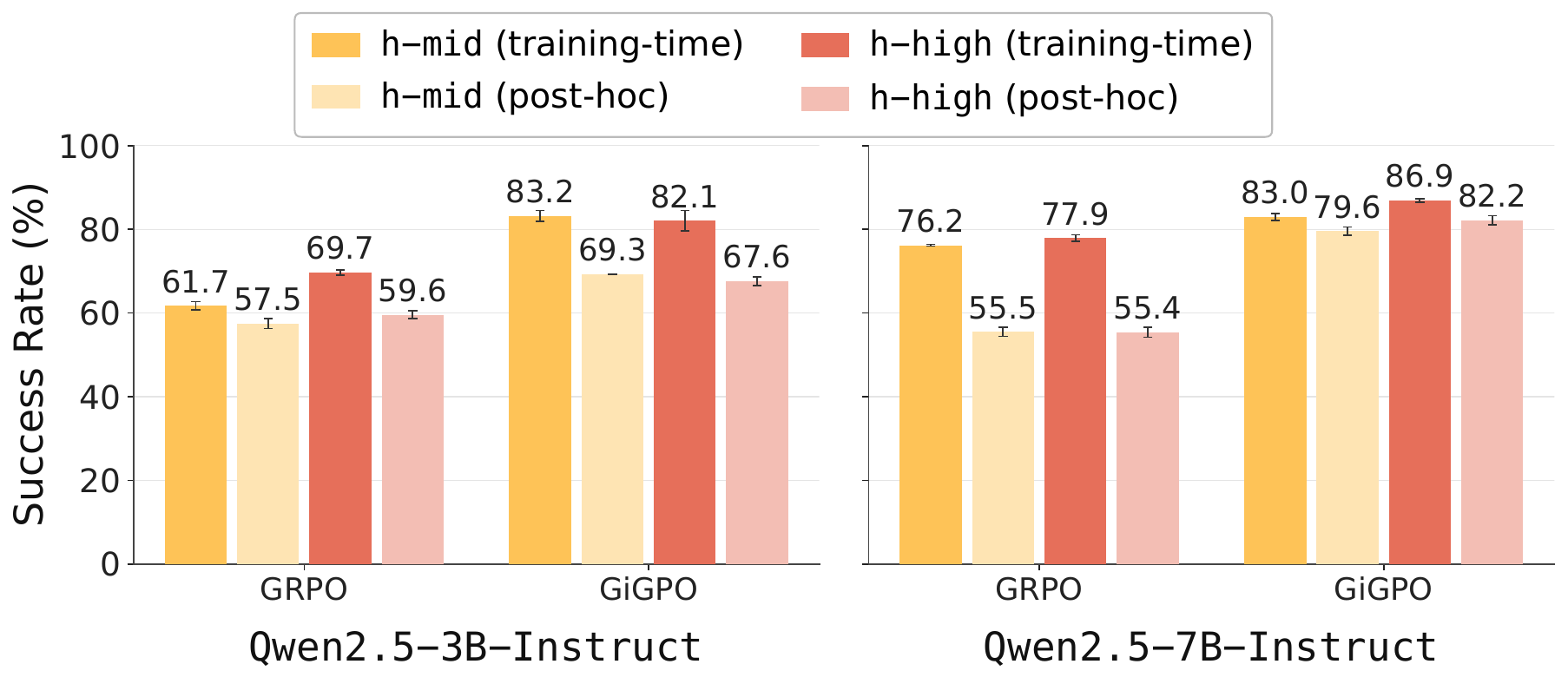}
    \caption{
        Training-time vs. post-hoc harness application: in-distribution success rates on $\mathcal{D}^{\text{te}}_{\text{all}}$.
        For each harness $\{ \texttt{h-mid}, \texttt{h-high} \}$, we compare an agent post-trained with that harness in place (\emph{training-time}) against an agent post-trained under \texttt{h-low} with that harness applied only at evaluation (\emph{post-hoc}). Training-time application consistently outperforms post-hoc application.
    }
    \label{fig:posthoc}
\end{figure}

\subsection{Post-hoc Harness Application (RQ2)}
\label{sec:iid_posthoc}
\begin{observation}[Post-hoc Gap]
    \label{obs:post-hoc}
    Applying a harness only after training recovers little of the benefit of training with it in place.
\end{observation}

Once the harness is treated as a controllable variable, a natural question arises: should {\color{black} an informative harness} be applied at training time, or in a post-hoc manner after post-training under a low-effort harness?
{\color{black} For closed-source models, post-hoc harness engineering is the default practice \cite{lee2026meta}, so one might expect the same recipe to suffice for open-source models after post-training as well. Our results show that it does not: we}
find that harness-aware post-training is preferable across all model and harness configurations (\cref{fig:posthoc}).
The gap is particularly large for \texttt{Qwen2.5-7B-Instruct} post-trained with GRPO, where training-time harness application outperforms post-hoc application by $20.7$ points under \texttt{h-mid} and $22.5$ points under \texttt{h-high}.

{\color{black} In addition, we conduct a short adaptation experiment: 
Starting from the model checkpoint post-trained with GiGPO under $\texttt{h-low}$ ($\texttt{Qwen2.5-3B-Instruct}$), we continue post-training under $\texttt{h-mid}$ or $\texttt{h-high}$ using $20\%$ of the main training budget.
Short adaptation improves over naive post-hoc application, from 69.3 to 71.5 under $\texttt{h-mid}$ and from 67.6 to 75.9 under $\texttt{h-high}$, but agents post-trained with the target harness from the outset remain the strongest, leaving gaps of 11.7 and 6.2 points.
The same trend holds for GRPO (Table~\ref{tab:posthoc_adaptation})}. Short adaptation thus partially closes the post-hoc gap, while early alignment between the harness and post-training remains preferable.
This suggests that the harness is better specified before post-training, allowing the agent to adapt to the harness it will ultimately use.

\subsection{OOD Robustness ({\color{black}RQ3})}
\label{sec:ood}
\begin{figure}[t]
    \centering
    \includegraphics[width=\linewidth]{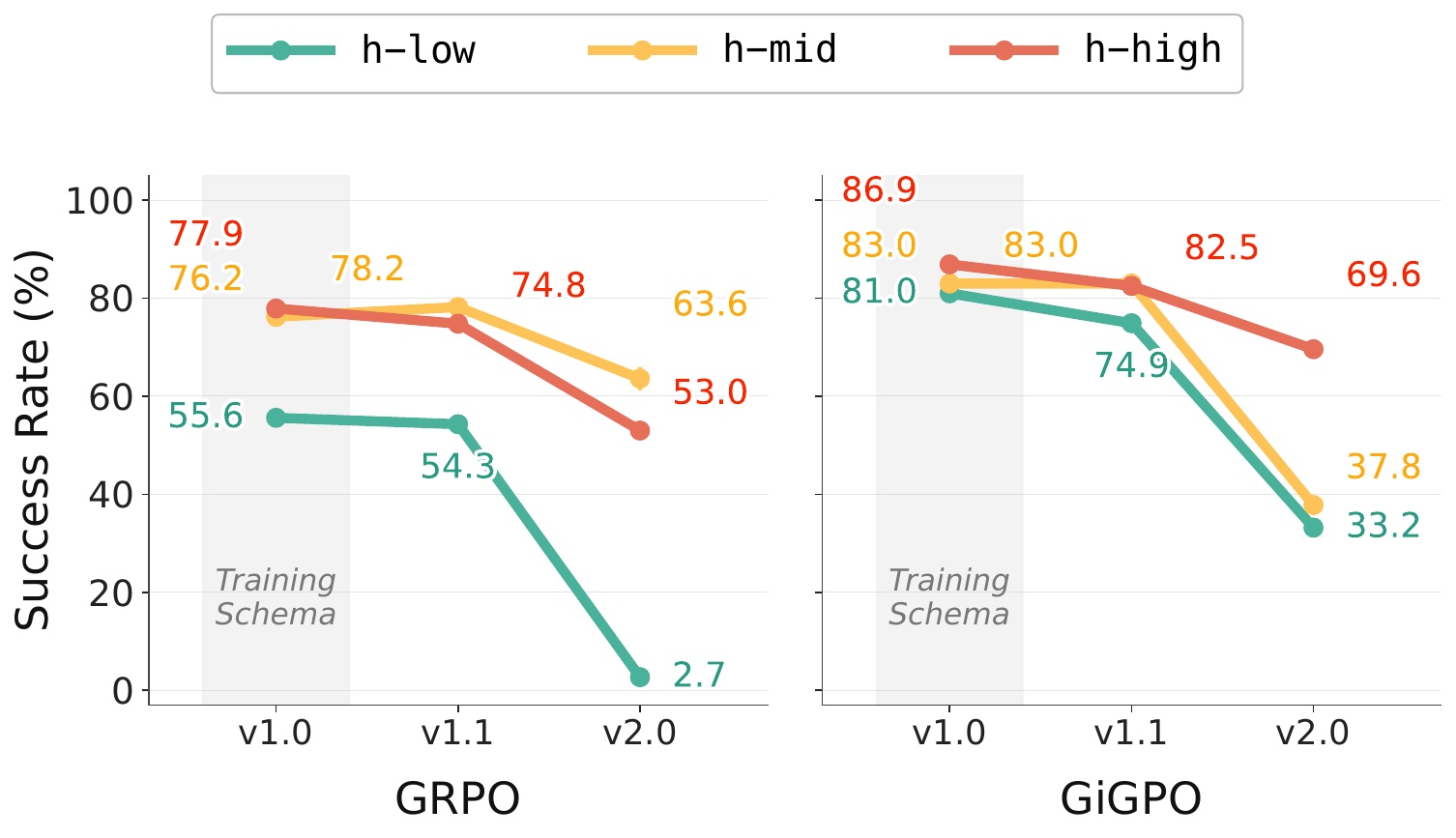}
    \caption{
        Post-training success rates of \texttt{Qwen2.5-\allowbreak 7B-\allowbreak Instruct} under tool environment shift:
        Agents are post-trained on $\mathcal{D}^{\text{tr}}_{\text{all}}$ under each harness with tool schema fixed to \texttt{v1.0}, and evaluated on $\mathcal{D}_{\text{all}}^{\text{te}}$ under the same harness but with schema \texttt{v1.1} (mild shift) or \texttt{v2.0} (stronger shift).
        Across both algorithms, more informative harnesses {\color{black} generally} yield greater robustness to such shifts.
    }
    \label{fig:tes_7b}
\end{figure}


\begin{observation}[Tool Environment Shift]
    \label{obs:tes}
    Harness-aware post-training mitigates the performance drop under {tool environment shift}, while post-training under \texttt{h-low} with \textcolor{black}{low design} effort suffers a drastic drop under stronger shift.
\end{observation}

{\color{black}\begin{table*}[t]
  \centering
  
  \footnotesize
  \setlength{\tabcolsep}{4pt}
  \renewcommand{\arraystretch}{1.25}
  \resizebox{\textwidth}{!}{%
  \begin{tabular}{@{}c l l l@{}}
    \toprule
    \multicolumn{1}{c}{\textbf{Harness}} 
    & \multicolumn{1}{c}{\textbf{Training-time Form (\texttt{v1.0})}} 
    & \multicolumn{1}{c}{\textbf{Test-time Form (\texttt{v2.0})}} 
    & \multicolumn{1}{c}{\textbf{Agent Output}} \\
    \midrule
    \multirow{3}{*}{\texttt{h-low}}
      & \texttt{Go(receptacle=``countertop 1'')}    & \texttt{ReceptacleControl(action=``navigate\_to'', target=``countertop 1'')}  & \texttt{GoTo(``countertop 1'')} \\
      & \texttt{Go(receptacle=``shelf 1'')}         & \texttt{ReceptacleControl(action=``navigate\_to'', target=``shelf 1'')}       & \texttt{GoToLocation(``shelf 1'')} \\
      & \texttt{Go(receptacle=``diningtable 1'')}   & \texttt{ReceptacleControl(action=``navigate\_to'', target=``diningtable 1'')} & \texttt{GoToDiningTable(``diningtable 1'')} \\
    \midrule
    \multirow{3}{*}{\texttt{h-mid}}
      & \texttt{Go(receptacle=``countertop 1'')}    & \texttt{ReceptacleControl(action=``navigate\_to'', target=``countertop 1'')}  & \texttt{navigate\_to(``countertop 1'')} \\
      & \texttt{Go(receptacle=``dresser 1'')}       & \texttt{ReceptacleControl(action=``navigate\_to'', target=``dresser 1'')}     & \texttt{go\_to(``dresser 1'')} \\
      & \texttt{Go(receptacle=``dresser 1'')}       & \texttt{ReceptacleControl(action=``navigate\_to'', target=``dresser 1'')}     & \texttt{goto\_dresser()} \\
    \bottomrule
  \end{tabular}}%
  \caption{Example failure modes of \texttt{Qwen2.5-7B-Instruct} post-trained with GiGPO under \texttt{h-low} and \texttt{h-mid}, with tool schema \texttt{v1.0} at training time and \texttt{v2.0} at evaluation. The corresponding valid forms under \texttt{v1.0} and \texttt{v2.0} are shown for reference. The agents fail to adapt to the test-time schema, generating plausible-looking but non-existent tool names.}
  \label{tab:v2_failure_examples}
\end{table*}}
Harness-aware post-training {\color{black} not only tends to improve} in-distribution performance (Obs.~\ref{obs:iid}), but also {\color{black} mitigates the performance drop} under tool environment shift (\cref{fig:tes_7b}).
While the agent post-trained {\color{black}under a harness with low design effort} (\texttt{h-low}) adapts to the mild shift (\texttt{v1.1}), it suffers a drastic performance drop under \texttt{v2.0}.
In particular, \texttt{Qwen2.5-7B-Instruct} post-trained with GRPO under \texttt{h-low} achieves $2.7$ under \texttt{v2.0}, which is $10.8$ points below the base model without post-training ($13.5$, \cref{tab:exp1_toolshift_extended}).

{\color{black}

\begin{figure}[t]
  \centering
  \includegraphics[width=\linewidth]{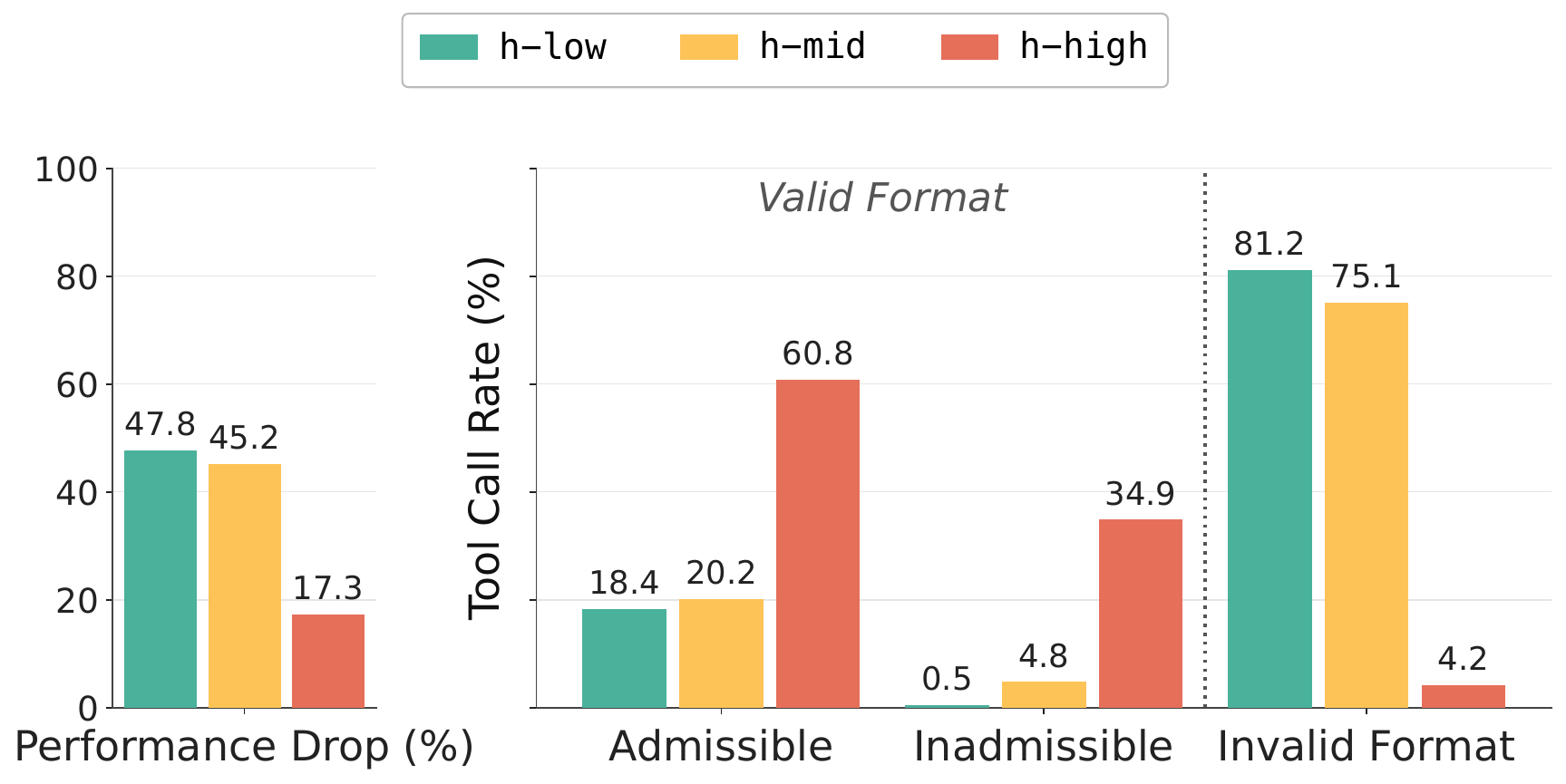}
    \caption{In-depth analysis of post-training under the tool shift from \texttt{v1.0} (training) to \texttt{v2.0} (evaluation).
    (Left) Performance drop under \texttt{v2.0} relative to that of in-distribution.
    (Right) Breakdown of generated tool calls into three categories: \emph{admissible} (correct format and executable), \emph{inadmissible} (correct format but rejected by the environment, \ie \texttt{``Nothing happens''}), and \emph{invalid} (malformed, \eg non-existent tool name or argument errors).}
  \label{fig:tes_ablation}
\end{figure}

While \texttt{Qwen2.5-7B-Instruct} post-trained with GiGPO under \texttt{h-low} performs well in-distribution (\cref{fig:iid}), it exhibits a substantial drop under \texttt{v2.0} (\cref{fig:tes_7b}), indicating that the agent fails to adapt to the test-time schema.
As such, we conduct an in-depth analysis on the tool calling patterns of \texttt{Qwen2.5-7B-Instruct} post-trained with GiGPO, where tool schema is fixed to \texttt{v1.0} at training time and \texttt{v2.0} at evaluation.

The model post-trained under \texttt{h-high} generates $95.7\%$ valid tool calls under \texttt{v2.0}, indicating that it adapts well to the tool environment shift at the surface level (\cref{fig:tes_ablation}).
However, $34.9\%$ of these calls are not admissible at the current state, resulting in a performance drop of $17.3$ points (from $86.9$ to $69.6$) relative to its in-distribution result.
By contrast, the models post-trained under \texttt{h-mid} and \texttt{h-low} fail to adapt at the schema level, returning $e_t = \texttt{``Invalid tool format''}$ in $75.1\%$ and $81.2\%$ of attempts, respectively.
Performance degrades by $45.2$ and $47.8$ points under \texttt{v2.0}, significantly larger than that of \texttt{h-high}.
This underscores the importance of training with an informative harness for OOD robustness, particularly under stronger tool environment shift.
Representative failure modes are provided in \cref{tab:v2_failure_examples}.
}

\begin{figure}[t]
    \centering
    \includegraphics[width=\linewidth]{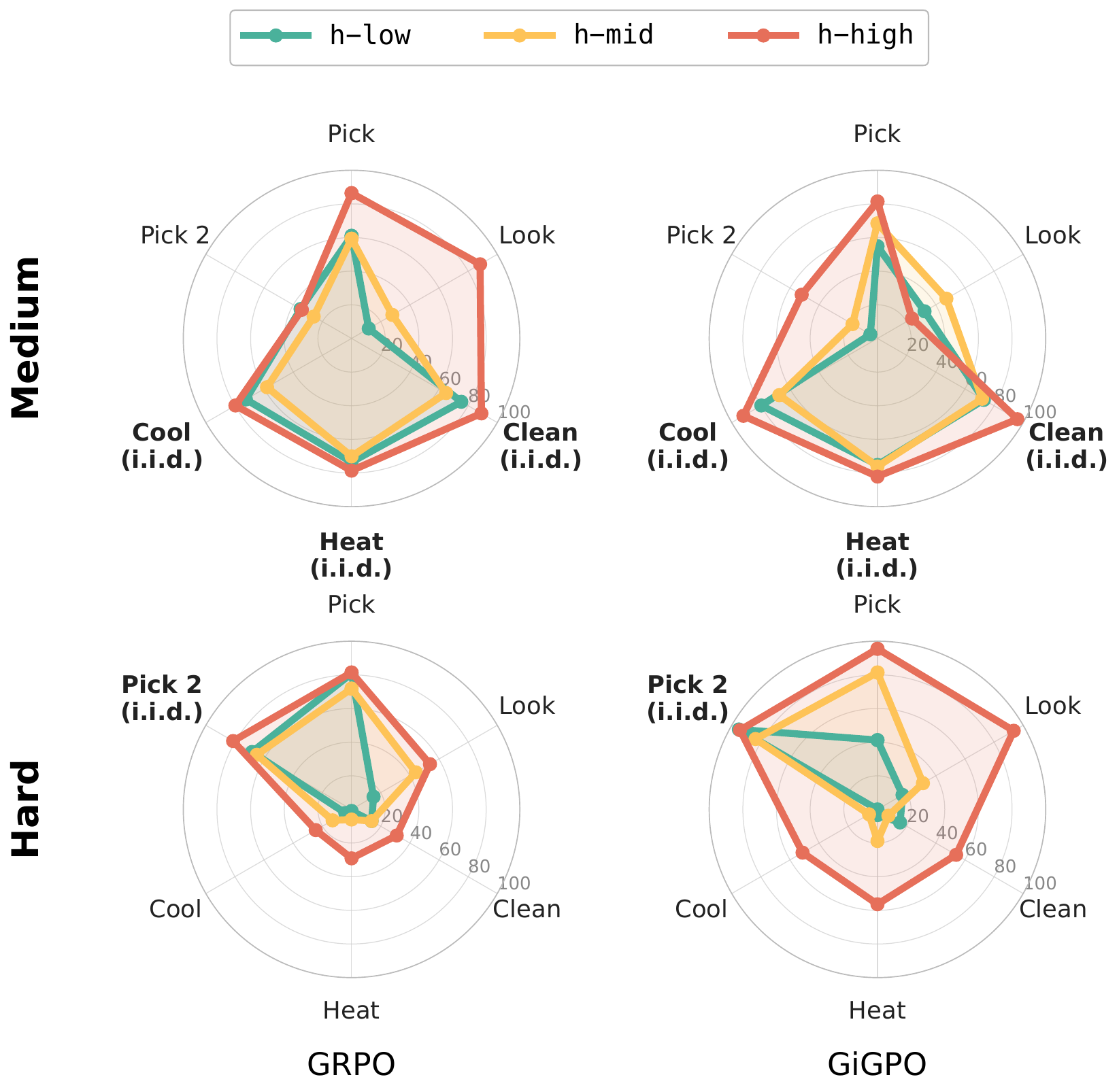}
    \caption{
        Post-training success rates of \texttt{Qwen2.5-\allowbreak 7B-\allowbreak Instruct} under task shift: agents are post-trained on $\mathcal{D}^{\text{tr}}_{\text{med}}$ (top row) and $\mathcal{D}^{\text{tr}}_{\text{hard}}$ (bottom row) under each harness with tool schema fixed to \texttt{v1.0}, and evaluated on $\mathcal{D}^{\text{te}}_{\text{all}}$ under the same harness and schema. In-distribution task categories are highlighted in \textbf{bold}: \textbf{\texttt{Clean}}, \textbf{\texttt{Heat}}, \textbf{\texttt{Cool}} for $\mathcal{D}^{\text{tr}}_{\text{med}}$, and \textbf{\texttt{Pick\,2}} for $\mathcal{D}^{\text{tr}}_{\text{hard}}$. Post-training under informative harnesses tends to improve performance on OOD task categories.
    }
    \label{fig:ts_7b}
\end{figure}

\begin{observation}[Task Shift]
    \label{obs:ts}
    Prior knowledge encoded in the harness boosts inter-task transfer of agent performance.
\end{observation}
Performance {\color{black} generally improves} with harness informativeness in the task shift scenario as well, where this trend holds across both algorithms (\cref{fig:ts_7b}).
GRPO improves from $33.1$ under \texttt{h-low} to $50.6$ under \texttt{h-high} ($+17.5$), and GiGPO from $28.8$ to $73.4$ ($+44.6$, \cref{tab:iid_posttraining_hard}).
This improvement arises not only from the in-distribution task category (\texttt{Pick\,2}), but also from the OOD categories, where agents post-trained under \texttt{h-low} fail to transfer (\cref{fig:ts_7b}).

Specifically, \texttt{Qwen2.5-7B-Instruct} post-trained with GRPO under \texttt{h-low} achieves $0.9$ on \texttt{Heat}, which rises to $29.1$ under \texttt{h-high} (\cref{tab:iid_posttraining_hard}).
A similar pattern holds with GiGPO under \texttt{h-low}, which achieves $0.0$ on \texttt{Cool} and rises to $51.4$ under \texttt{h-high}.
This indicates that the prior knowledge encoded in an informative harness helps the agent generalize across task groups.
{\color{black} Similar trends are observed in additional experiments under fine-grained task shift settings, in which agents are post-trained on a single task category (Table~\ref{tab:single_category_taskshift}).}
As in the tool environment shift scenario, harness-aware post-training improves OOD robustness.

\section{Conclusion}
\label{sec:conclusion}
In this paper, we studied harness design in the context of post-training tool-integrated LLM agents, a factor that prior work has either treated as a fixed engineering detail or examined only in the training-free regime.
Building on \texttt{ALFWorld}, we constructed a benchmark that treats the harness as a controllable design dimension and supports four evaluation regimes: zero-shot, and post-training under the in-distribution and two OOD scenarios (tool environment shift and task shift).
Across these regimes, we instantiated three harnesses with varying levels of informativeness and analyzed how harness design influences post-training of LLM agents in both in-distribution and OOD settings.

Our experiments establish that harness design and post-training cannot be treated as separable design choices.
Performance improves monotonically with harness informativeness at zero-shot, and this trend {\color{black} largely carries over} to post-training under the in-distribution scenario.
This improvement, however, requires the harness to be in place during post-training: applying it only afterward recovers little of the benefit of training with one in place.
Under OOD settings, harness-aware post-training mitigates the performance drop under tool environment shift and helps the agent generalize across task groups, whereas post-training under a harness with low design effort suffers a drastic performance drop under stronger tool environment shift and fails to transfer to OOD task categories.
These results demonstrate that harness-aware post-training is a prerequisite for, rather than a supplement to, robust performance of tool-integrated LLM agents.


\section*{Limitations}
\label{sec:limitations}

Our {\color{black} main} analysis is conducted on a single environment, \texttt{ALFWorld}.
Although our benchmark built on \texttt{ALFWorld} enables a controlled study of harness design under in-distribution and OOD settings, validating our findings on benchmarks with different task distributions and tool environments is an important direction.
{\color{black} To partially address concerns about the generalizability of our findings, we replicate our tool environment shift protocol on \texttt{WebShop} \cite{yao2022webshop}, an interactive e-commerce benchmark.
The central pattern reproduces: the agent post-trained under the more informative harness consistently outperforms its low-effort counterpart under the shifted tool schema with both GRPO and GiGPO, and the agent post-trained under the low-effort harness collapses under GiGPO (success rate $54.7 \rightarrow 33.5$) despite reasonable in-distribution performance (Table~\ref{tab:rebuttal_toolshift_score_turns}).} 

We examine harness-aware post-training with two open-source models, \texttt{Qwen2.5-3B-Instruct} and \texttt{Qwen2.5-7B-Instruct}, and two RL algorithms, GRPO and GiGPO, spanning $24$ training configurations per model and amounting to approximately $1{,}800$ H200 GPU-hours for the main \texttt{ALFWorld} experiments.
Extending this to a broader range of models and RL algorithms, beyond what our compute budget allowed, is left to future work.

We instantiate three harnesses (\texttt{h-low}, \texttt{h-mid}, \texttt{h-high}) as representative levels of informativeness.
In practice, informative harnesses such as \texttt{h-mid} and \texttt{h-high} must be obtained either through expert design or costly exploration of the environment (\S\ref{sec:harness}).
While GiGPO under \texttt{h-low} performs comparably to harness-aware counterparts under the in-distribution scenario, it still breaks down under stronger tool environment shift, indicating that finer credit assignment alone cannot replace an informative harness.
A natural direction is therefore to develop harness-aware post-training methods that jointly optimize the harness and the LLM agent.
Our benchmark, which supports arbitrary harness designs, can serve as a testbed for such algorithms.

{\color{black} As a final remark, the scope of our findings is confined to small-to-medium-scale open-source LLMs post-trained for bounded, domain-specific tool-use workflows. We emphasize that this regime is of growing practical importance, particularly in terms of cost efficiency: agentic workloads issue large numbers of repetitive tool calls, so a compact model post-trained for a specific workflow is substantially more economical to serve than a frontier model invoked at every step, while also meeting latency, privacy, and on-device constraints. This trend is illustrated by $\texttt{FunctionGemma}$, a 270M-parameter open model designed for customized local function calling \cite{black2025functiongemma}, and by $\texttt{xLAM}$, a family of action models from Salesforce whose compact variants are post-trained specifically for function calling in agentic systems \cite{zhang2025xlam}. In such deployments, the harness and the post-trained model are co-designed for a fixed target workflow, which is the setting our study directly informs.}

\section*{Acknowledgements}
This work was supported by the National Research Foundation of Korea (NRF) grant funded by the Korea government (MSIT) (RS-2024-00337955); by the Institute of Information \& Communications Technology Planning \& Evaluation (IITP) grants funded by the Korea government (MSIT) (RS-2019-II191906, Artificial Intelligence Graduate School Program (POSTECH); RS-2024-00457882, National AI Research Lab Project and Artificial Intelligence Research Hub Project; RS-2024-00509258 and RS-2024-00469482, Global AI Frontier Lab; RS-2025-00560062); and by IITP under the Leading Generative AI Human Resources Development (IITP-2026-RS-2026-25546560) grant funded by the Korea government (MSIT).
\bibliography{llm}

\clearpage
\crefalias{section}{appendix}
\crefalias{subsection}{appendix}
\appendix


\section{Related Work}
\label{sec:related}

\subsection{Training Tool-integrated Agentic Systems}
A substantial body of prior work aims to improve the performance of tool-integrated agentic systems under static tool environments.
Most approaches train an LLM agent to generate valid tool calling sequences using supervised fine-tuning (SFT) with expert demonstrations or reinforcement learning (RL).

\citet{zhang2025agent} propose two SFT strategies that incorporate additional trajectories generated by the agent itself.
The first performs ``implicit world modeling'' to learn transition dynamics from agent--environment interactions before fine-tuning on expert trajectories.
The second constructs ``self-reflection data'' by contrasting agent-generated trajectories with expert demonstrations, encouraging the model to internalize why expert solutions are preferable.
Similarly, \citet{qian2025toolrl} design reward functions that align generated trajectories with expert demonstrations, supervising both intermediate tool calls and final outcomes.

In parallel, a complementary line of work develops reinforcement learning methods tailored to tool-integrated agents~\citep{jin2025search, mai2025agent, xue2025simpletir, jiang-etal-2025-s3, mo2025multi, cheng2025agent}.
A key distinction from conventional RL is the masking of feedback tokens from the tool environment, ensuring that parameter updates depend solely on agent-generated tokens.
Additional refinements include filtering trajectories containing invalid or incomplete tool calls~\citep{xue2025simpletir} and introducing multi-turn optimization strategies, such as turn-level ratios in group relative policy optimization (GRPO).
Beyond individual algorithms, prior studies examine the interaction between SFT and RL for improving generalization~\citep{chu2025sft}: SFT followed by RL can enhance training stability, while RL-only approaches have been reported to achieve stronger generalization in certain tool-integrated reasoning tasks~\citep{xue2025simpletir}.

Despite their methodological differences, these approaches share two assumptions: that the tool environment is static throughout training and evaluation, and that the harness is fixed and treated as part of the environment rather than as a design variable.
As a consequence, neither the harness's role in shaping what the post-trained agent learns, nor the agent's robustness to changes in the harness or the tool environment, is systematically examined.

\subsection{Harness Design for Tool-integrated Agentic Systems}
The harness, as introduced in \S\ref{sec:intro}, is widely recognized as a substantial determinant of agent performance, with recent work reporting accuracy gaps of up to roughly six-fold across harnesses~\citep{lee2026meta}.
For tool-integrated agentic systems, the harness specifies, among other things, which tools are exposed to the agent and how they are described, what auxiliary information accompanies each per-step observation, and how the resulting context is assembled into a prompt.

In practice, harnesses have been hand-designed for closed-source models.
Benchmarks for interactive LLM agents, including those built on $\texttt{ALFWorld}$, typically report results obtained under carefully crafted harnesses, often containing detailed descriptions of available actions, structured per-step observations, and various forms of auxiliary information~\citep{zhang2025agent}.
A recent line of work seeks to remove the human from this loop: \citet{lee2026meta} introduce Meta-Harness, which casts harness design as a search problem and uses a high-capacity LLM agent to read candidate harnesses and propose new ones automatically.
Even with such automated search and strong base models, however, the gains from harness design alone are not always satisfactory.
Furthermore, the underlying LLM is held fixed throughout, with the implications for harness design in the post-training regime remaining unaddressed.

In this work, we study how harness design and post-training jointly shape the resulting agent: we treat the harness as an explicit, controlled design dimension and post-train independently under harnesses providing three different levels of auxiliary information, characterizing their interplay both in-distribution and under task and tool environment shift.

\subsection{Tool-integrated Agentic Systems under Dynamic Environments}
\citet{chen2025toolevo} introduce ToolQA-D, a benchmark in which API names, signatures, and behaviors drift over time with the agent unaware of these changes, and propose ToolEVO, a framework that uses Monte Carlo Tree Search to enable an agent to detect and adapt to such variability at inference time.
More recently, \citet{li2026proevolve} propose ProEvolve, a graph-based framework in which the data, tools, and schema of an agent environment are represented as a typed relational graph and evolved through programmable graph transformations, allowing controlled evaluation of agents' robustness as environments change.
While these efforts share our motivation, they neither treat the harness as a design dimension nor consider the post-training regime, leaving the interplay between the two under such shifts uncharacterized.

\clearpage


\onecolumn

\section{Benchmark Details}
\label{app:benchmark_details}

\subsection{Harness Configurations}
\label{app:harness_examples}

\begin{center}
\tiny
\renewcommand{\arraystretch}{1.2}
\setlength{\tabcolsep}{6pt}
\setlength{\LTcapwidth}{\linewidth}

\begin{longtable}{
    c
    m{0.30\linewidth}
    m{0.45\linewidth}
}
\label{tab:tool_desc_comparison} \\
\toprule
\textbf{Tool} & \multicolumn{1}{c}{\textbf{\texttt{h-low} / \texttt{h-mid}}} & \multicolumn{1}{c}{\textbf{\texttt{h-high}}} \\
\midrule
\endfirsthead

\toprule
\textbf{Tool} & \multicolumn{1}{c}{\textbf{\texttt{h-low} / \texttt{h-mid}}} & \multicolumn{1}{c}{\textbf{\texttt{h-high}}} \\
\midrule
\endhead

\midrule
\endfoot

\bottomrule \\
\caption{One-line versus rich tool descriptions in $p$, for all 13 tools (under tool schema \texttt{v1.0}). The one-line form is shared by \texttt{h-low} and \texttt{h-mid}; the rich form is used only by \texttt{h-high}. Rows are grouped according to the \texttt{v2.0} consolidation (Table~\ref{tab:tool_schema_consolidation_v20}).} \\
\endlastfoot

\texttt{Go}        & \texttt{Move to a receptacle.} & \texttt{Move to a receptacle in the room. This sets your current location to that receptacle; you must be at a receptacle before you can Take from it, Open/Close it, Examine it, or Move an object to it.} \\
\addlinespace
\texttt{Open}      & \texttt{Open a receptacle.} & \texttt{Open a closed container at your current location (drawer, cabinet, fridge, microwave). You must be at the container (use Go) and it must currently be closed. After opening, its contents become visible and you can Take items from it.} \\
\addlinespace
\texttt{Close}     & \texttt{Close a receptacle.} & \texttt{Close an open container at your current location. Normally not required for task completion; useful only when a goal explicitly requires closed state.} \\
\addlinespace
\texttt{Take}      & \texttt{Take an object from a receptacle.} & \texttt{Pick up an object from a receptacle at your current location. Preconditions: you must be at the receptacle (use Go), the object must be visible, and if the receptacle is a closed container (drawer/cabinet/fridge/microwave) you must Open it first. You can only carry one object at a time.} \\
\addlinespace
\texttt{Move}      & \texttt{Place an object in or on a receptacle.} & \texttt{Place the object you are carrying in/on a receptacle at your current location. Preconditions: you must be carrying the object and be at the target receptacle (for closed containers, Open it first). This action COMPLETES the task when the carried object and the target receptacle both match the goal.} \\
\addlinespace
\texttt{Clean}     & \texttt{Clean an object using a receptacle.} & \texttt{Clean an object you are carrying using a sinkbasin at your current location. Preconditions: you must be carrying the object and be at the sinkbasin.} \\
\addlinespace
\texttt{Heat}      & \texttt{Heat an object using a receptacle.} & \texttt{Heat an object you are carrying using a microwave at your current location. Preconditions: you must be carrying the object and be at the microwave.} \\
\addlinespace
\texttt{Cool}      & \texttt{Cool an object using a receptacle.} & \texttt{Cool an object you are carrying using a fridge at your current location. Preconditions: you must be carrying the object and be at the fridge.} \\
\addlinespace
\texttt{Slice}     & \texttt{Slice an object using a sharp object.} & \texttt{Slice an object at your current location using a knife (or butterknife) you are carrying. Preconditions: you must be carrying the sharp object and be at the target object's location.} \\
\addlinespace
\texttt{Look}      & \texttt{Look around your current location.} & \texttt{Look around your current surroundings. If not at any receptacle, reports all visible receptacles in the room. If at a receptacle, reports what is on/in it.} \\
\addlinespace
\texttt{Inventory} & \texttt{Check your current inventory.} & \texttt{Report what object you are currently carrying. You can carry at most one object at a time.} \\
\addlinespace
\texttt{Examine}   & \texttt{Examine a receptacle or an object.} & \texttt{Examine a receptacle or object at your current location. For containers this reveals their contents; for objects this reports their state. Useful for re-checking an already-opened receptacle without Go-ing again.} \\
\addlinespace
\texttt{Use}       & \texttt{Use an object.} & \texttt{Activate an object at your current location, most commonly turning on a lamp (desklamp / floorlamp). In `look at object in light' tasks, turning on the lamp while holding the target object completes the task.} \\
\end{longtable}
\end{center}

\clearpage

\begin{table}[h]
\centering
\small
\renewcommand{\arraystretch}{1.3}
\setlength{\tabcolsep}{6pt}
\begin{tabular}{
    c
    >{\ttfamily}m{0.78\linewidth}
}
\toprule
\textbf{Harness Ver.} & \multicolumn{1}{c}{\textbf{Per-step history $\mathcal{T}_t$}} \\
\midrule
\texttt{h-low}
    & Your task is to: <GOAL>. \newline
      Current location: <LOC> \newline
      You are now at step <N> and your current observation is: <STEP\_OBS> \\
\addlinespace
\texttt{h-mid}
    & Your task is to: <GOAL>. \newline
      Current location: <LOC> \newline
      You are now at step <N> and your current observation is: <STEP\_OBS> \newline
      Valid tools: <TOOL\_A>, <TOOL\_B>, <TOOL\_C> \\
\addlinespace
\texttt{h-high}
    & Your task is to: <GOAL>. \newline
      Current location: <LOC> \newline
      Carrying: <ITEM\_OR\_"nothing"> \newline
      You are now at step <N> and your current observation is: <STEP\_OBS> \newline
      Valid tools: <TOOL\_A>, <TOOL\_B>, <TOOL\_C> \\
\bottomrule
\end{tabular}
\label{tab:perstep_comparison}
\caption{Per-step history $\mathcal{T}_t$ under each harness version. The three harnesses share a common core, with each richer version appending one additional auxiliary line (\texttt{Valid tools:} under \texttt{h-mid}, and additionally \texttt{Carrying:} under \texttt{h-high}).}
\end{table}

\clearpage

\subsection{Tool Schema Versions}
\label{app:tool_schema_details}

\begin{table}[hbt!]
\centering
\small
\renewcommand{\arraystretch}{1.1}
\setlength{\tabcolsep}{8pt}
\begin{tabular}{c c c}
\toprule
\textbf{Meaning} & \textbf{\texttt{v1.0}} & \textbf{\texttt{v1.1}} \\
\midrule
Move to a receptacle                 & \texttt{Go}        & \texttt{NavigateTo}     \\
Open a container                     & \texttt{Open}      & \texttt{OpenContainer}  \\
Close a container                    & \texttt{Close}     & \texttt{CloseContainer} \\
\addlinespace
Pick up an object                    & \texttt{Take}      & \texttt{Pickup}         \\
Place an object in/on a receptacle   & \texttt{Move}      & \texttt{Place}          \\
\addlinespace
Wash an object (in sinkbasin)        & \texttt{Clean}     & \texttt{Wash}           \\
Heat an object (in microwave)        & \texttt{Heat}      & \texttt{Warm}           \\
Cool an object (in fridge)           & \texttt{Cool}      & \texttt{Chill}          \\
Slice an object (with a knife)       & \texttt{Slice}     & \texttt{Cut}            \\
\addlinespace
Survey the surroundings              & \texttt{Look}      & \texttt{SurveyRoom}     \\
Check items currently carried        & \texttt{Inventory} & \texttt{CheckCarried}   \\
Examine a receptacle or object       & \texttt{Examine}   & \texttt{Inspect}        \\
\addlinespace
Activate a device (e.g., lamp)       & \texttt{Use}       & \texttt{Activate}       \\
\bottomrule
\end{tabular}
\caption{Tool name mapping from \texttt{v1.0} to \texttt{v1.1}, covering all 13 tools defined for \texttt{ALFWorld}. Rows are grouped according to the \texttt{v2.0} consolidation (Table~\ref{tab:tool_schema_consolidation_v20}).}
\label{tab:tool_schema_name_mapping_v11}
\end{table}

\begin{table}[hbt!]
\centering
\small
\renewcommand{\arraystretch}{1.1}
\setlength{\tabcolsep}{8pt}
\begin{tabular}{c c c}
\toprule
\textbf{Meaning} & \textbf{\texttt{v1.0}} & \textbf{\texttt{v1.1}} \\
\midrule
Destination of a navigation/placement action       & \texttt{receptacle} & \texttt{destination} \\
Object being acted upon                            & \texttt{object}     & \texttt{item}        \\
Source receptacle (e.g., for \texttt{Take})        & \texttt{receptacle} & \texttt{source}      \\
Washing receptacle (\texttt{sinkbasin})            & \texttt{receptacle} & \texttt{washer}      \\
Heating receptacle (\texttt{microwave})            & \texttt{receptacle} & \texttt{heater}      \\
Cooling receptacle (\texttt{fridge})               & \texttt{receptacle} & \texttt{cooler}      \\
Slicing instrument                                 & \texttt{knife}      & \texttt{blade}       \\
Target of an examination                           & \texttt{target}     & \texttt{subject}     \\
\bottomrule
\end{tabular}
\label{tab:tool_schema_param_mapping_v11}
\caption{Parameter key mapping from \texttt{v1.0} to \texttt{v1.1}, covering all parameter roles shared across tools.}
\end{table}

\clearpage

\begin{table}[h]
\centering
\small
\renewcommand{\arraystretch}{1.3}
\setlength{\tabcolsep}{6pt}
\begin{tabular}{
    >{\centering\arraybackslash}m{3.0cm}
    >{\centering\arraybackslash}m{3.2cm}
    >{\centering\arraybackslash}m{3.2cm}
    >{\centering\arraybackslash}m{2.4cm}
}
\toprule
\textbf{\texttt{v2.0} Tool} & \textbf{Merged \texttt{v1.1} Tools} & \textbf{\texttt{action} Values} & \textbf{Unified Parameters} \\
\midrule
\texttt{ReceptacleControl}
    & \texttt{NavigateTo}, \texttt{OpenContainer}, \texttt{CloseContainer}
    & \texttt{navigate\_to}, \texttt{open\_container}, \texttt{close\_container}
    & \texttt{action}, \texttt{target} \\
\addlinespace
\texttt{ObjectTransport}
    & \texttt{Pickup}, \texttt{Place}
    & \texttt{pickup}, \texttt{place}
    & \texttt{action}, \texttt{item}, \texttt{location} \\
\addlinespace
\texttt{ObjectTransform}
    & \texttt{Wash}, \texttt{Warm}, \texttt{Chill}, \texttt{Cut}
    & \texttt{wash}, \texttt{warm}, \texttt{chill}, \texttt{cut}
    & \texttt{action}, \texttt{item}, \texttt{instrument} \\
\addlinespace
\texttt{Observe}
    & \texttt{SurveyRoom}, \texttt{CheckCarried}, \texttt{Inspect}
    & \texttt{survey\_room}, \texttt{check\_carried}, \texttt{inspect}
    & \texttt{action}, \texttt{subject} \\
\addlinespace
\texttt{Activate}
    & \texttt{Activate}
    & -- (single operation)
    & \texttt{device} \\
\bottomrule
\end{tabular}
\caption{Tool consolidation from \texttt{v1.1} (13 tools) to \texttt{v2.0} (5 tools). Each \texttt{v2.0} group merges one or more \texttt{v1.1} tools sharing structural or functional similarity, distinguished within the group by a discrete \texttt{action} parameter. Parameter keys are unified across merged operations.}
\label{tab:tool_schema_consolidation_v20}
\end{table}

\clearpage


\twocolumn

{\color{black}\section{Experimental Details}
\label{app.exp_details}

\paragraph{Multi-step Tool Calling Process as Markov Decision Process.} We illustrate in detail how multi-step tool calling process is modeled as a Markov decision process (MDP) at a sequence level.
Specifically, the MDP is defined by a tuple $
    \mathcal{M}
    \coloneqq 
    (
        \mathcal{S}, \mathcal{A}, \mathcal{P}, r, \mathcal{P}_0
    )
$ as the following.
\begin{itemize}[leftmargin=*]
    \item $\mathcal{S} \coloneqq \mathcal{V}^\ast$: State space

    \item $\mathcal{A} \coloneqq \mathcal{V}^\ast$: Action space

    \item $\mathcal{P}: \mathcal{S} \times \mathcal{A} \times \mathcal{S} \rightarrow [0, 1]$: State transition probability such that $
        \sum_{s' \in \mathcal{S}}
        \mathcal{P}(s, a, s') = 1
    $ for all $s \in \mathcal{S}$ and $a \in \mathcal{A}$.
    In addition, we let $
        \mathcal{P}(
            s' | s, a
        )
        \coloneqq 
        \mathcal{P}(
            s, a, s'
        )
    $.

     \item $
        r: 
        \mathcal{S} \times \mathcal{A} \times \mathcal{S} 
        \rightarrow
        \mathbb{R}
    $: Step-level reward function evaluated at each step, where we let
    \begin{align*}
        r_t 
        \coloneqq 
        r( s_t, a_t, s_{t+1} ).
    \end{align*}

    \item $
        \mathcal{P}_0: \mathcal{S} \rightarrow [0, 1]
    $: Initial state probability distribution
\end{itemize}
Then, the multi-step tool calling process can be formalized as follows.
Before the process starts, the initial state $s_1 \coloneqq (p, q)$ is sampled from $\mathcal{P}_0$, where $p \in \mathcal{V}^\ast$ is a system prompt and $q \in \mathcal{V}^\ast$ is a user query.
In each tool calling step $t$, the LLM agent $\pi_\theta: \mathcal{S} \rightarrow \mathcal{A}$ first observes the state $s_t$.
Specifically, $s_t \coloneqq ( p,  q, \mathcal{T}_{t-1} ) \in \mathcal{S}$, where $\mathcal{T}_{t-1}$ is a history of tool calling steps up to the $(t-1)$-th step (possibly in a summarized version), and $\mathcal{T}_0$ is an empty sequence.
The agent then generates an action sequence $a_t \in \mathcal{A}$.
The environment then samples the next state $s_{t+1}$ from the probability distribution $
    \mathcal{P}(\cdot | s_t, a_t),
$
where
\begin{align*}
    \mathcal{P}( s_{t+1} | s_t, a_t )
    \coloneqq
    \begin{cases}
        1 & s_{t+1} = (p, q, \mathcal{T}_t), \\
        0 & \text{o.w.}
    \end{cases}
\end{align*}
Here, $
    e_t\!=\!\texttt{TE}((s_t, a_t))
$ and $
    \mathcal{T}_t\!=\!\texttt{TE}((\mathcal{T}_{t-1}, a_t, e_t)).
$
Note that the transition dynamics is deterministic and \texttt{TE}-dependent.
In addition, the agent receives the step level reward $
    r_t
$.
The tool calling process proceeds for $T \in \mathbb{N}$ steps, where the  process terminates either when it succeeds in a given task or when the agent reaches the maximum number of tool calling steps.
Finally, by letting $
    \tau
    \coloneqq
    \{
        ( s_1, a_1, s_2 ),
        \dots, 
        ( s_T, a_T, s_{T+1} )
    \}
    \subset 
    (\mathcal{S} \times \mathcal{A} \times \mathcal{S})^\ast
$ be an episode-level trajectory and $
    \mathcal{R}: (\mathcal{S} \times \mathcal{A} \times \mathcal{S})^\ast
    \rightarrow
    \mathbb{R}
$ be an episode-level reward, the objective can be formulated as 
\begin{align*}
    \mathcal{J} (\theta)
    = 
    \mathbb{E}_{\mathcal{P}_0, \mathcal{P}, \pi_\theta} \left[ 
        \mathcal{R}( \tau )        
    \right],
\end{align*}
where $
    \mathcal{R} ( \tau )
    \coloneqq
    \sum_{t=1}^T \gamma^{t-1} r_t
$
is a typical choice ($\gamma \in [0, 1]$).}

\paragraph{Group-relative policy optimization (GRPO) \citep{shao2024deepseekmath}.} It is a widely adopted algorithm for post-training LLMs due to its computational efficiency in calculating the gradient $\nabla_\theta \mathcal{J}(\theta)$.
Given a system prompt and a question pair $(p, q)$, a behavior policy $\pi_{\theta_{\text{old}}}$ samples a group of $G$ episodes $\{\tau^i\}_{i=1}^G$, where $
    \tau^i
    \coloneqq 
    \{
        ( s_1^i, a_1^i, s_2^i ),
        \dots,
        ( s_{T^i}^i, a_{T^i}^i, s_{T^i+1}^i )
    \}
$ is the $i$-th episode.
Here, we let $
    a_t^i
    \coloneqq 
    ( a_{t1}^i, \dots, a_{tn_t^i}^i ),
$ where $n_t^i \coloneqq | a_t^i |$.
GRPO algorithm then optimizes the following token-level objective:
\begin{equation*}
\begin{aligned}
\mathcal{J}(\theta)
&=
\mathbbm{E}_{\substack{
{\scriptscriptstyle
(p, q) \sim P_0,}\\
\scriptscriptstyle\{ {\color{black} a_t^i} \}
\sim {\pi_{\theta_{\text{old}}}}
}}
\Bigg[
\frac{1}{G}
\sum_{i=1}^G
\frac{1}{\sum_{t=1}^{T^i}n_t^i}
\sum_{t=1}^{T^i} 
\sum_{k=1}^{n_t^i}
\\
&~
\bigg\{
\text{min}\left( 
{\color{black} \rho_{tk}^i} {\color{black} \hat{A}^i},
\text{clip}\left( 
{\color{black} \rho_{tk}^i},
1 - \epsilon, 
1 + \epsilon
\right)
{\color{black} \hat{A}^i}
\right)
\\
&~~
-
\beta 
D_{\text{KL}} \left( 
\pi_\theta( \cdot \mid s_{<tk}^i ) \, \| \, 
\pi_{\text{ref}}( \cdot \mid s_{<tk}^i )
\right)
\bigg\}
\Bigg].
\end{aligned}
\end{equation*}
where $s_{<tk}^i \coloneqq ( p, q, \mathcal{T}_{t-1}^i, a_{t1}^i, \dots, a_{t(k-1)}^i )$ 
denotes the context preceding token $a_{tk}^i$, and the token-level ratio $\rho_{tk}^i$ 
and the episode-level advantage $\hat{A}^i$ are defined as follows:
\begin{equation*}
        \begin{aligned}
            \rho_{tk}^i
            = 
            \frac{
                \pi_{\theta} ( a_{tk}^i \mid s_{<tk}^i )
            }{
                \pi_{\theta_{\text{old}}} ( a_{tk}^i \mid s_{<tk}^i )        
            },
            \quad 
            \hat{A}^i 
            =
            \frac{R^i - \mu}{\sigma}.    
        \end{aligned}
\end{equation*}
Here, $\{ R^i \}_{i=1}^G$ are episode-level rewards evaluated on $\{ \tau^i \}_{i=1}^G$, 
and 
$
    \mu = \texttt{mean}( \{ R^i \}_{i=1}^G )
$, $
    \sigma = \texttt{std}( \{ R^i \}_{i=1}^G )
$ are their sample mean and standard deviation, respectively.
Here, $\pi_{\text{ref}}$ is a fixed reference policy, typically initialized from the 
initial policy and kept frozen throughout training. In practice, the per-token KL 
term is computed using the low-variance estimator of $D_{\text{KL}}(\pi_\theta \| \pi_{\text{ref}})$, as adopted in GRPO \citep{shao2024deepseekmath}.

\paragraph{Group-in-group policy optimization (GiGPO) \citep{feng2026groupingroup}.}
Unlike GRPO, GiGPO differentiates the advantage assigned to each step within the same episode by additionally considering a {\color{black} step-level advantage $\hat{A}_t^i$} in addition to an {\color{black} episode-level advantage $\hat{A}^i$}.
To define a step-level advantage, GiGPO algorithm requires a {\color{black} state identifier $
    h: \mathcal{V}^\ast \times \mathcal{V}^\ast \rightarrow \{ 0, 1 \}
$} that verifies whether two different per-step initial states, $s_{t_1}^i$ and $s_{t_2}^j$, \ie the states from which steps $t_1$ and $t_2$ begin, are semantically equivalent (possibly from different episodes, where $i \neq j$).
Specifically, given a set of step-level transitions across all $G$ episodes
$
    \mathcal{G}
    \coloneqq 
    \left\{ 
        ( 
            s_{t}^i, a_t^i, s_{t+1}^i
        )
    \right\}_{\scriptscriptstyle i \in [G], t \in [T^i]},
$
{\color{black} GiGPO partitions $\mathcal{G} = \cup_{\tilde{s} \in \mathcal{AN}} ~ \mathcal{G}( \tilde{s} )$ into disjoint groups}, 
where steps within each group $\mathcal{G} ( \tilde{s} )$ shares the same per-step initial state $\tilde{s} \in \mathcal{V}^\ast$ w.r.t. the state identifier $h$ as follows:
\begin{align*}
    \mathcal{G} ( \tilde{s} )
    \coloneqq 
    \left\{ 
        ( s_{t}^i, a_t^i, s_{t+1}^i ) \in \mathcal{G}
        : 
        h( s_{t}^i, \tilde{s} ) = 1
    \right\}.
\end{align*}
Here, $\mathcal{AN}$ is an {\color{black} anchor set} that partitions $\mathcal{G}$ based on the semantic equivalence of the per-step initial state that jointly satisfies
\begin{equation*}
    \begin{aligned}
        \mathcal{G} 
        &= 
        \cup_{\tilde{s} \in \mathcal{AN}} \mathcal{G} ( \tilde{s} ),
        \\
        \mathcal{G} ( \tilde{s} ) \cap \mathcal{G} ( \hat{s} ) 
        = 
        \emptyset &~ \forall \tilde{s}, \hat{s} \in \mathcal{AN} ~ \text{s.t.} ~ h ( \tilde{s}, \hat{s} ) = 0.
    \end{aligned}
\end{equation*}
Then, every step level transition $( s_{t}^i, a_t^i, s_{t+1}^i )$ is assigned an index $g_t^i$ that specifies which group that it belongs to, defined as $
    g_t^i 
    \coloneqq 
    ( \tilde{s} \in \mathcal{AN} ~ \text{s.t.} ~ h ( s_{t}^i, \tilde{s} ) = 1 ).
$
GiGPO algorithm then optimizes the following token-level objective:
\begin{equation*}
\begin{aligned}
\mathcal{J}(\theta)
&=
\mathbbm{E}_{\substack{
{\scriptscriptstyle
(p, q) \sim P_0,}\\
\scriptscriptstyle\{ {\color{black} a_t^i} \}
\sim {\pi_{\theta_{\text{old}}}}
}}
\Bigg[
\frac{1}{G}
\sum_{i=1}^G
\frac{1}{\sum_{t=1}^{T^i}n_t^i}
\sum_{t=1}^{T^i} 
\sum_{k=1}^{n_t^i}
\\
&~
\bigg\{
\text{min}\Big( 
{\color{black} \rho_{tk}^i} {\color{black} ( \hat{A}^i + \omega \hat{A}_t^i)},
\\
&\qquad
\text{clip}\left( 
{\color{black} \rho_{tk}^i},
1 - \epsilon, 
1 + \epsilon
\right)
{\color{black} ( \hat{A}^i + \omega \hat{A}_t^i)}
\Big)
\\
&~~
-
\beta 
D_{\text{KL}} \left( 
\pi_\theta( \cdot \mid s_{<tk}^i ) \, \| \, 
\pi_{\text{ref}}( \cdot \mid s_{<tk}^i )
\right)
\bigg\}
\Bigg].
\end{aligned}
\end{equation*}
where the token level ratio $\rho_{tk}^i$ and the episode-level advantage $\hat{A}^i$ are defined in a same manner as GRPO, and the step-level advantage $\hat{A}^i_t$ is defined as :
\begin{align*}
    \hat{A}_t^i 
    = 
    \frac{
        R_t^i - \mu_{g_t^i}
    }{\sigma_{g_t^i}}.   
\end{align*}
Here, $
    R_t^i
    \coloneqq 
    \sum_{\tilde{t} = t}^{T^i} \gamma^{\tilde{t} - t} r_{\tilde{t}}^i
$ is a step-level return of the $t$-th step within the $i$-th episode.
For each anchor $\tilde{s} \in \mathcal{AN}$, $\mu_{\tilde{s}}$ and $\sigma_{\tilde{s}}$ denote the sample mean and standard deviation of the step-level returns $\{ R_t^i \}$ within $\mathcal{G}(\tilde{s})$.
The advantage $\hat{A}_t^i$ then uses the statistics of the group $g_t^i$ to which the $t$-th step of the $i$-th episode belongs.
Note that the discount factor $\gamma \in [0, 1]$ may be different from that used to define the episode-level reward $R^i$.
$\omega \in \mathbbm{R}^+$ is a hyperparameter that balances episode-level and step-level advantages.


\paragraph{Training.}
We fine-tune Qwen2.5-3B-Instruct and Qwen2.5-7B-Instruct with GRPO and GiGPO, using group-normalized advantages over $G{=}8$ trajectories per prompt. We train for 500 steps with a constant learning rate of $1{\times}10^{-6}$, a batch of 16 task prompts per step, a PPO minibatch size of 256, and a low-variance KL penalty (coefficient $0.01$) applied to the loss. A randomly sampled 10\% of the training data is held out for validation and used to select the final checkpoint. Rollouts use vLLM with a sampling temperature of $1.0$ and a maximum of 50 turns per episode over the 13 ALFWorld tools, with a maximum prompt length of 3072 tokens and a maximum response length of 1024 tokens. The reward is sparse: a successful trajectory receives $10$, while each invalid action incurs a penalty of $-0.1$. We use $4 \times$ H200 GPUs for training.

\paragraph{Evaluation.}
We evaluate on the ALFWorld test split of 274 games (140 seen, 134 unseen), with the evaluation harness matched to the training harness and a sampling temperature of $0.4$. Each configuration is run with three seeds, and we report the mean success rate; the corresponding per-configuration standard deviations are reported alongside the means in all result tables. For GPT-5-mini, we use a reasoning effort of \texttt{high}; since reasoning models require substantially more tokens, we set the maximum response length to 4096 tokens. Due to budget constraints, GPT-5-mini is run with a single seed. For the tool environment shift scenario, we refer to the paraphrased schema \texttt{v1.1} as the \emph{mild} shift and the consolidated schema \texttt{v2.0} as the \emph{strong} shift, reflecting the increasing degree of surface-level deviation from the \texttt{v1.0} training schema.

\clearpage
\onecolumn

\section{Additional Experimental Results}
\label{app:add_experiments}


\begin{table}[!htbp]
\centering
\resizebox{\textwidth}{!}{%
\begin{tabular}{c c c *{7}{c}}
\toprule
\multirow{2}{*}[-1ex]{\textbf{Model Type}} & \multirow{2}{*}[-1ex]{\textbf{Model}} &
\multirow{2}{*}[-1ex]{\textbf{Test-time Harness}}
& \multicolumn{2}{c}{$\mathcal{D}^{\text{te}}_{\text{easy}}$}
& \multicolumn{3}{c}{$\mathcal{D}^{\text{te}}_{\text{med}}$}
& \multicolumn{1}{c}{$\mathcal{D}^{\text{te}}_{\text{hard}}$}
& \multirow{2}{*}[-1ex]{All} \\
\cmidrule(lr){4-5} \cmidrule(lr){6-8} \cmidrule(lr){9-9}
& &
& Pick & Look & Clean & Heat & Cool & Pick 2 & \\
\midrule
\multirow{3}{*}{Closed-source}
& \multirow{3}{*}{GPT-5 Mini (high)}
& \texttt{h-low} & $71.2$ & $16.1$ & $19.0$ & $12.8$ & $15.2$ & $17.1$ & $28.1$ \\
&& \texttt{h-mid} & $67.8$ & $12.9$ & $20.7$ & $20.5$ & $34.8$ & $12.2$ & $31.0 \, (+2.9)$ \\
&& \texttt{h-high} & $93.2$ & $77.4$ & $36.2$ & $66.7$ & $78.3$ & $61.0$ & $68.3 \, (+40.2)$ \\
\midrule
\multirow{6}{*}{Open-source}
& \multirow{3}{*}{Qwen2.5-3B-Instruct}
& \texttt{h-low} & $6.2_{\pm 2.6}$ & $0.0_{\pm 0.0}$ & $0.0_{\pm 0.0}$ & $0.0_{\pm 0.0}$ & $0.0_{\pm 0.0}$ & $0.0_{\pm 0.0}$ & $1.3_{\pm 0.6}$ \\
&& \texttt{h-mid} & $11.9_{\pm 4.5}$ & $2.2_{\pm 1.9}$ & $0.0_{\pm 0.0}$ & $3.4_{\pm 3.0}$ & $0.0_{\pm 0.0}$ & $0.0_{\pm 0.0}$ & $3.3_{\pm 1.1} \, (+2.0)$ \\
&& \texttt{h-high} & $35.0_{\pm 2.6}$ & $15.1_{\pm 1.9}$ & $11.5_{\pm 1.0}$ & $12.8_{\pm 0.0}$ & $2.2_{\pm 2.2}$ & $0.0_{\pm 0.0}$ & $13.9_{\pm 0.7} \, (+12.6)$ \\
\cmidrule(lr){2-10}
& \multirow{3}{*}{Qwen2.5-7B-Instruct}
& \texttt{h-low} & $19.8_{\pm 3.5}$ & $18.3_{\pm 4.9}$ & $0.6_{\pm 1.0}$ & $3.4_{\pm 3.0}$ & $2.9_{\pm 3.3}$ & $0.0_{\pm 0.0}$ & $7.4_{\pm 0.9}$ \\
&& \texttt{h-mid} & $44.6_{\pm 2.0}$ & $20.4_{\pm 3.7}$ & $9.8_{\pm 3.6}$ & $8.5_{\pm 3.9}$ & $5.1_{\pm 1.3}$ & $0.0_{\pm 0.0}$ & $16.1_{\pm 1.7} \, (+8.7)$ \\
&& \texttt{h-high} & $54.8_{\pm 3.9}$ & $18.3_{\pm 6.7}$ & $26.4_{\pm 2.6}$ & $35.9_{\pm 0.0}$ & $24.6_{\pm 3.3}$ & $1.6_{\pm 1.4}$ & $29.0_{\pm 1.8} \, (+21.6)$ \\
\bottomrule
\end{tabular}
}
\caption{
Zero-shot success rates without post-training: Pretrained agents are evaluated on $\mathcal{D}^{\text{te}}_{\text{all}}$ under each harness with tool schema fixed to \texttt{v1.0}.
GPT-5 Mini is run only once due to the limited budget.}
\label{tab:exp0_iid_zeroshot}
\end{table}

  \begin{table}[!htbp]
    \centering
    \resizebox{0.82\linewidth}{!}{
    \begin{tabular}{c c c *{7}{c}}
    \toprule
    \multirow{2}{*}[-1ex]{\textbf{Model}} & \multirow{2}{*}[-1ex]{\textbf{Algorithm}} &
    \multirow{2}{*}[-1ex]{\textbf{Harness}}
    & \multicolumn{2}{c}{$\mathcal{D}^{\text{te}}_{\text{easy}}$}
    & \multicolumn{3}{c}{$\mathcal{D}^{\text{te}}_{\text{med}}$}
    & \multicolumn{1}{c}{$\mathcal{D}^{\text{te}}_{\text{hard}}$}
    & \multirow{2}{*}[-1ex]{All} \\
    \cmidrule(lr){4-5} \cmidrule(lr){6-8} \cmidrule(lr){9-9}
    & & & Pick & Look & Clean & Heat & Cool & Pick 2 & \\
    \midrule
    \multirow{6}{*}{Qwen2.5-3B-Instruct}
    & \multirow{3}{*}{GRPO}
    & \texttt{h-low} & $80.2_{\pm 2.6}$ & $38.7_{\pm 3.2}$ & $66.1_{\pm 3.6}$ & $64.1_{\pm 0.0}$   & $66.7_{\pm 1.3}$ &   $0.0_{\pm 0.0}$  & ${\underline{56.0}}_{\pm 1.1}$ \\
    &                       & \texttt{h-mid}  & $80.2_{\pm 2.6}$ & $60.2_{\pm 4.9}$ & $67.2_{\pm 1.7}$ & $70.9_{\pm
  1.5}$  & $68.8_{\pm 4.5}$ & $11.4_{\pm 3.7}$ & ${\underline{61.7}}_{\pm 1.0}$ \\
    &                       & \texttt{h-high} & $81.4_{\pm 1.7}$ & $77.4_{\pm 5.6}$ & $75.9_{\pm 3.0}$ & $71.8_{\pm
  2.6}$  & $83.3_{\pm 1.3}$ & $21.1_{\pm 3.7}$ & ${\underline{69.7}}_{\pm 0.6}$ \\
    \cmidrule(lr){2-10}
    & \multirow{3}{*}{GiGPO} & \texttt{h-low} & $91.5_{\pm 0.0}$ & $89.2_{\pm 6.7}$ & $72.4_{\pm 7.5}$ & $67.5_{\pm
  3.0}$   & $64.5_{\pm 1.3}$ & $0.0_{\pm 0.0}$  & ${\underline{65.6}}_{\pm 1.3}$ \\
    && \texttt{h-mid}  & $94.9_{\pm 1.7}$ & $63.4_{\pm 4.9}$ & $93.7_{\pm 2.6}$ & $85.5_{\pm 3.0}$ & $82.6_{\pm 2.2}$ &
   $65.0_{\pm 6.1}$ & ${\underline{83.2}}_{\pm 1.3}$ \\
    &                       & \texttt{h-high} & $85.9_{\pm 5.4}$ & $88.2_{\pm 1.9}$ & $89.7_{\pm 1.7}$ & $84.6_{\pm
  5.1}$ & $79.7_{\pm 5.5}$ &  $61.8_{\pm 1.4}$ & ${\underline{82.1}}_{\pm 2.4}$ \\
    \midrule
    \multirow{6}{*}{Qwen2.5-7B-Instruct}
    & \multirow{3}{*}{GRPO}
    & \texttt{h-low} & $80.8_{\pm 2.0}$ & ${{0.0}}_{\pm 0.0}$ & $56.9_{\pm 1.7}$ &  $75.2_{\pm 3.0}$ &
  $52.9_{\pm 1.3}$ & $43.9_{\pm 2.4}$ & ${\underline{55.6}}_{\pm 0.8}$ \\
    &                       & \texttt{h-mid}  & $79.7_{\pm 1.7}$ & $86.0_{\pm 1.9}$ & $78.2_{\pm 1.0}$ & $71.8_{\pm
  2.6}$  & $66.7_{\pm 2.5}$ & $75.6_{\pm 2.4}$ & ${\underline{76.2}}_{\pm 0.2}$ \\
    &                       & \texttt{h-high} & $82.5_{\pm 2.0}$ & $86.0_{\pm 1.9}$ & $81.6_{\pm 1.0}$ & $75.2_{\pm
  1.5}$  & $79.7_{\pm 4.5}$ & $61.0_{\pm 4.2}$ & ${\underline{77.9}}_{\pm 0.8}$ \\
    \cmidrule(lr){2-10}
    & \multirow{3}{*}{GiGPO}
    & \texttt{h-low} & $86.4_{\pm 1.7}$ & $75.3_{\pm 3.7}$ & $75.9_{\pm 0.0}$ & $76.9_{\pm 0.0}$ & $89.9_{\pm 1.3}$ &
  $78.9_{\pm 2.8}$ & ${\underline{81.0}}_{\pm 0.0}$ \\
    && \texttt{h-mid}  & $90.4_{\pm 2.0}$ & $80.6_{\pm 6.5}$ & $87.4_{\pm 1.0}$ & $84.6_{\pm 0.0}$ & $72.5_{\pm 1.3}$ &
   $78.0_{\pm 2.4}$ & ${\underline{83.0}}_{\pm 0.8}$ \\
    && \texttt{h-high} & $89.3_{\pm 2.0}$ & $88.2_{\pm 1.9}$ & $92.5_{\pm 1.0}$ & $76.1_{\pm 1.5}$ & $86.2_{\pm 2.5}$ &
   $85.4_{\pm 2.4}$ & ${\underline{86.9}}_{\pm 0.4}$ \\
    \bottomrule
    \end{tabular}
    }
    \caption{
    In-distribution success rates after post-training:
    {\color{black}Agents are post-trained on $\mathcal{D}_{\text{all}}^{\text{tr}}$ under each harness with tool schema fixed to \texttt{v1.0}, and evaluated on $\mathcal{D}_{\text{all}}^{\text{te}}$ under the same harness and schema.
    }
    }
    \label{tab:exp0_iid_posttraining}
  \end{table}

\begin{table}[!htbp]
\centering
\resizebox{\linewidth}{!}{
\begin{tabular}{c c c *{7}{c}}
\toprule
\multirow{2}{*}[-1ex]{} & \multirow{2}{*}[-1ex]{\textbf{Algorithm}} & \multirow{2}{*}[-1ex]{\textbf{Post-hoc Harness}}  & \multicolumn{2}{c}{$\mathcal{D}^{\text{te}}_{\text{easy}}$}
& \multicolumn{3}{c}{$\mathcal{D}^{\text{te}}_{\text{med}}$}
& \multicolumn{1}{c}{$\mathcal{D}^{\text{te}}_{\text{hard}}$}
& \multirow{2}{*}[-1ex]{All} \\
\cmidrule(lr){4-5} \cmidrule(lr){6-8} \cmidrule(lr){9-9}
& & & Pick & Look & Clean & Heat & Cool & Pick 2 & \\
\midrule
\multirow{4}{*}{\resizebox{0.55em}{!}{\rotatebox{90}{Qwen2.5-3B-Instruct}}}
& \multirow{2}{*}{GRPO}
& \texttt{h-mid}  & $79.7_{\pm 3.4}$ & $49.5_{\pm 9.9}$ & $67.2_{\pm 3.0}$ & $63.2_{\pm 1.5}$ & $68.8_{\pm 1.3}$ &
$0.0_{\pm 0.0}$ & $57.5_{\pm 1.2} ~ ({\bf\underline{-4.2}})$ \\
& & \texttt{h-high} & $82.5_{\pm 1.0}$ & $50.5_{\pm 3.7}$ & $72.4_{\pm 1.7}$ & $68.4_{\pm 1.5}$ & $65.9_{\pm 1.3}$ &
$0.0_{\pm 0.0}$ & $59.6_{\pm 0.9} ~ ({\bf\underline{-10.1}})$ \\
\cmidrule(lr){2-10}
& \multirow{2}{*}{GiGPO}
& \texttt{h-mid}  & $89.8_{\pm 1.7}$ & $95.7_{\pm 1.9}$ & $77.0_{\pm 2.0}$ & $73.5_{\pm 1.5}$ & $73.9_{\pm 2.2}$ &
$0.0_{\pm 0.0}$ & $69.3_{\pm 0.0} ~ ({\bf\underline{-13.9}})$ \\
& & \texttt{h-high} & $89.3_{\pm 2.0}$ & $93.5_{\pm 3.2}$ & $73.6_{\pm 5.5}$ & $70.9_{\pm 1.5}$ & $72.5_{\pm 1.3}$ &
  $0.0_{\pm 0.0}$ & $67.6_{\pm 1.1} ~ ({\bf\underline{-14.5}})$ \\
\midrule
\multirow{4}{*}{\resizebox{0.55em}{!}{\rotatebox{90}{Qwen2.5-7B-Instruct}}}
& \multirow{2}{*}{GRPO}
& \texttt{h-mid}  & $77.4_{\pm 2.6}$ & $0.0_{\pm 0.0}$ & $56.9_{\pm 3.0}$ & $76.9_{\pm 4.4}$ & $56.5_{\pm 3.8}$ &
$42.3_{\pm 1.4}$ & $55.5_{\pm 1.1} ~ ({\bf\underline{-20.7}})$ \\
& & \texttt{h-high} & $80.2_{\pm 3.5}$ & $0.0_{\pm 0.0}$ & $58.0_{\pm 3.6}$ & $72.6_{\pm 1.5}$ & $55.8_{\pm 2.5}$ &
$40.7_{\pm 1.4}$ & $55.4_{\pm 1.2} ~ ({\bf\underline{-22.5}})$ \\
\cmidrule(lr){2-10}
& \multirow{2}{*}{GiGPO}
& \texttt{h-mid}  & $85.9_{\pm 1.0}$ & $55.9_{\pm 7.4}$ & $75.3_{\pm 1.0}$ & $82.9_{\pm 1.5}$ & $91.3_{\pm 0.0}$ &
$78.0_{\pm 2.4}$ & $79.6_{\pm 1.0} ~ ({\bf\underline{-3.4}})$ \\
& & \texttt{h-high} & $86.4_{\pm 0.0}$ & $80.6_{\pm 5.6}$ & $75.9_{\pm 0.0}$ & $79.5_{\pm 2.6}$ & $91.3_{\pm 0.0}$ &
$78.9_{\pm 1.4}$ & $82.2_{\pm 1.1} ~ ({\bf\underline{-4.7}})$ \\
\bottomrule
\end{tabular}}
\caption{
Training-time vs. post-hoc harness application: in-distribution success rates on $\mathcal{D}^{\text{te}}_{\text{all}}$.
For each harness $\{ \texttt{h-mid}, \texttt{h-high} \}$, we compare an agent post-trained with that harness in place (\emph{training-time}) against an agent post-trained under \texttt{h-low} with that harness applied only at evaluation (\emph{post-hoc}). 
The parenthesized value is the change in overall accuracy relative to post-training under the same harness applied throughout.
Training-time application consistently outperforms post-hoc application.
}
\label{tab:exp0_iid_posthoc}
\end{table}

\clearpage

\begin{table}[hbt!]
\centering
\resizebox{\textwidth}{!}{%
\begin{tabular}{c c c *{21}{c}}
\toprule
\multirow{3}{*}[-2ex]{} &
\multirow{3}{*}[-2ex]{} &
\multirow{3}{*}[-2ex]{\textbf{Harness}}
& \multicolumn{7}{c}{\textbf{v1.0}}
& \multicolumn{7}{c}{\textbf{v1.1}}
& \multicolumn{7}{c}{\textbf{v2.0}} \\
\cmidrule(lr){4-10} \cmidrule(lr){11-17} \cmidrule(lr){18-24}
& &
& \multicolumn{2}{c}{$\mathcal{D}^{\text{te}}_{\text{easy}}$}
& \multicolumn{3}{c}{$\mathcal{D}^{\text{te}}_{\text{med}}$}
& \multicolumn{1}{c}{$\mathcal{D}^{\text{te}}_{\text{hard}}$}
& \multirow{2}{*}[-1ex]{All}
& \multicolumn{2}{c}{$\mathcal{D}^{\text{te}}_{\text{easy}}$}
& \multicolumn{3}{c}{$\mathcal{D}^{\text{te}}_{\text{med}}$}
& \multicolumn{1}{c}{$\mathcal{D}^{\text{te}}_{\text{hard}}$}
& \multirow{2}{*}[-1ex]{All}
& \multicolumn{2}{c}{$\mathcal{D}^{\text{te}}_{\text{easy}}$}
& \multicolumn{3}{c}{$\mathcal{D}^{\text{te}}_{\text{med}}$}
& \multicolumn{1}{c}{$\mathcal{D}^{\text{te}}_{\text{hard}}$}
& \multirow{2}{*}[-1ex]{All} \\
\cmidrule(lr){4-5} \cmidrule(lr){6-8} \cmidrule(lr){9-9}
\cmidrule(lr){11-12} \cmidrule(lr){13-15} \cmidrule(lr){16-16}
\cmidrule(lr){18-19} \cmidrule(lr){20-22} \cmidrule(lr){23-23}
& &
& Pick & Look & Clean & Heat & Cool & Pick 2 &
& Pick & Look & Clean & Heat & Cool & Pick 2 &
& Pick & Look & Clean & Heat & Cool & Pick 2 & \\
\midrule
\multirow{10}{*}{\rotatebox{90}{{Qwen2.5-3B-Instruct}}}
& \multirow{3}{*}{\rotatebox{90}{\resizebox{3.5em}{!}{Zero-shot}}}
& \texttt{h-low} & $6.2$ & $0.0$ & $0.0$ & $0.0$ & $0.0$ & $0.0$ & ${\underline{1.3}}$ & $26.0$ & $14.0$ & $0.0$ & $0.0$ & $0.0$ & $0.0$ & ${\underline{7.2}}$ & $18.6$ & $10.8$ & $0.0$ & $0.0$ & $0.7$ & $0.0$ & ${\underline{5.4}}$ \\
&& \texttt{h-mid}  & $11.9$ & $2.2$ & $0.0$ & $3.4$ & $0.0$ & $0.0$ & ${\underline{3.3}}$ & $46.3$ & $25.8$ & $8.0$ & $4.3$ & $8.0$ & $0.0$ & ${\underline{16.5}}$ & $14.1$ & $18.3$ & $0.0$ & $0.0$ & $0.7$ & $0.0$ & ${\underline{5.2}}$ \\
&& \texttt{h-high} & $35.0$ & $15.1$ & $11.5$ & $12.8$ & $2.2$ & $0.0$ & ${\underline{13.9}}$ & $40.7$ & $20.4$ & $10.9$ & $14.5$ & $8.0$ & $0.0$ & ${\underline{16.8}}$ & $35.6$ & $28.0$ & $5.7$ & $7.7$ & $8.7$ & $0.0$ & ${\underline{14.6}}$ \\
\cmidrule(lr){2-24}
& \multirow{3}{*}{\rotatebox{90}{{GRPO}}}
& \texttt{h-low} & $80.2$ & $38.7$ & $66.1$ & $64.1$ & $66.7$ & $0.0$ & ${\underline{56.0}}$ & $52.0$ & $28.0$ & $10.9$ & $14.5$ & $5.8$ & $0.0$ & ${\underline{19.7}}$ & $17.5$ & $7.5$ & $0.6$ & $0.9$ & $2.2$ & $0.0$ & ${\underline{5.2}}$ \\
&& \texttt{h-mid}  & $80.2$ & $60.2$ & $67.2$ & $70.9$ & $68.8$ & $11.4$ & ${\underline{61.7}}$ & $72.3$ & $47.3$ & $56.3$ & $53.8$ & $62.3$ & $14.6$ & ${\underline{53.2}}$ & $48.0$ & $26.9$ & $12.1$ & $20.5$ & $26.1$ & $7.3$ & ${\underline{24.3}}$ \\
&& \texttt{h-high} & $81.4$ & $77.4$ & $75.9$ & $71.8$ & $83.3$ & $21.1$ & ${\underline{69.7}}$ & $71.2$ & $63.4$ & $70.1$ & $70.9$ & $84.8$ & $26.0$ & ${\underline{65.6}}$ & $73.4$ & $58.1$ & $66.1$ & $51.3$ & $79.0$ & $26.8$ & ${\underline{60.9}}$ \\
  \cmidrule(lr){2-24}
  & \multirow{3}{*}{\rotatebox{90}{{GiGPO}}}
  & \texttt{h-low} & $91.5$ & $89.2$ & $72.4$ & $67.5$ & $64.5$ & $0.0$ & ${\underline{65.6}}$ & $87.0$ & $93.5$ &
  $69.5$ & $68.4$ & $57.2$ & $0.0$ & ${\underline{63.4}}$ & $20.9$ & $90.3$ & $27.0$ & $27.4$ & $38.4$ & $0.0$ &
  ${\underline{30.8}}$ \\ 
  && \texttt{h-mid} & $94.9$ & $63.4$ & $93.7$ & $85.5$ & $82.6$ & $65.0$ & ${\underline{83.2}}$ & $92.7$ & $45.2$ &
  $87.9$ & $74.4$ & $83.3$ & $54.5$ & ${\underline{76.4}}$ & $89.8$ & $37.6$ & $47.7$ & $35.0$ & $63.8$ & $47.2$ &
  ${\underline{56.4}}$ \\ 
  && \texttt{h-high} & $85.9$ & $88.2$ & $89.7$ & $84.6$ & $79.7$ & $61.8$ & ${\underline{82.1}}$ & $81.9$ & $83.9$ &
  $82.2$ & $87.2$ & $81.2$ & $63.4$ & ${\underline{80.0}}$ & $85.9$ & $71.0$ & $61.5$ & $64.1$ & $79.7$ & $60.2$ &
  ${\underline{71.0}}$ \\ 
\midrule
\multirow{10}{*}{\rotatebox{90}{{Qwen2.5-7B-Instruct}}}
& \multirow{3}{*}{\rotatebox{90}{\resizebox{3.5em}{!}{Zero-shot}}}
& \texttt{h-low} & $19.8$ & $18.3$ & $0.6$ & $3.4$ & $2.9$ & $0.0$ & ${\underline{7.4}}$ & $15.8$ & $25.8$ & $5.7$ & $0.9$ & $1.4$ & $0.0$ & ${\underline{7.9}}$ & $35.6$ & $24.7$ & $5.7$ & $3.4$ & $6.5$ & $1.6$ & ${\underline{13.5}}$ \\
&& \texttt{h-mid}  & $44.6$ & $20.4$ & $9.8$ & $8.5$ & $5.1$ & $0.0$ & ${\underline{16.1}}$ & $58.8$ & $22.6$ & $35.1$ & $20.5$ & $22.5$ & $6.5$ & ${\underline{30.3}}$ & $58.8$ & $32.3$ & $16.1$ & $18.8$ & $17.4$ & $8.9$ & ${\underline{26.6}}$ \\
&& \texttt{h-high} & $54.8$ & $18.3$ & $26.4$ & $35.9$ & $24.6$ & $1.6$ & ${\underline{29.0}}$ & $39.5$ & $16.1$ & $27.0$ & $35.9$ & $22.5$ & $8.9$ & ${\underline{26.3}}$ & $54.8$ & $46.2$ & $31.6$ & $35.9$ & $26.8$ & $16.3$ & ${\underline{35.8}}$ \\
\cmidrule(lr){2-24}
& \multirow{3}{*}{\rotatebox{90}{{GRPO}}}
& \texttt{h-low} & $80.8$ & $0.0$ & $56.9$ & $75.2$ & $52.9$ & $43.9$ & ${\underline{55.6}}$ & $77.4$ & $0.0$ & $59.2$ & $63.2$ & $55.1$ & $45.5$ & ${\underline{54.3}}$ & $2.3$ & $0.0$ & $8.0$ & $0.9$ & $0.7$ & $1.6$ & ${\underline{2.7}}$ \\
&& \texttt{h-mid}  & $79.7$ & $86.0$ & $78.2$ & $71.8$ & $66.7$ & $75.6$ & ${\underline{76.2}}$ & $83.6$ & $79.6$ & $87.4$ & $76.9$ & $63.8$ & $74.0$ & ${\underline{78.2}}$ & $78.0$ & $77.4$ & $73.0$ & $29.1$ & $52.9$ & $64.2$ & ${\underline{63.6}}$ \\
&& \texttt{h-high} & $82.5$ & $86.0$ & $81.6$ & $75.2$ & $79.7$ & $61.0$ & ${\underline{77.9}}$ & $78.0$ & $83.9$ & $73.6$ & $76.9$ & $76.1$ & $61.8$ & ${\underline{74.8}}$ & $66.7$ & $76.3$ & $28.7$ & $48.7$ & $51.4$ & $56.1$ & ${\underline{53.0}}$ \\
  \cmidrule(lr){2-24}
  & \multirow{3}{*}{\rotatebox{90}{{GiGPO}}}
  & \texttt{h-low} & $86.4$ & $75.3$ & $75.9$ & $76.9$ & $89.9$ & $78.9$ & ${\underline{81.0}}$ & $87.6$ & $21.5$ &
  $75.9$ & $73.5$ & $90.6$ & $79.7$ & ${\underline{74.9}}$ & $36.2$ & $0.0$ & $61.5$ & $30.8$ & $40.6$ & $8.1$ &
  ${\underline{33.2}}$ \\ 
  && \texttt{h-mid} & $90.4$ & $80.6$ & $87.4$ & $84.6$ & $72.5$ & $78.0$ & ${\underline{83.0}}$ & $92.7$ & $82.8$ &
  $87.4$ & $82.9$ & $69.6$ & $78.0$ & ${\underline{83.0}}$ & $32.2$ & $24.7$ & $52.3$ & $52.1$ & $41.3$ & $17.9$ &
  ${\underline{37.8}}$ \\ 
  && \texttt{h-high} & $89.3$ & $88.2$ & $92.5$ & $76.1$ & $86.2$ & $85.4$ & ${\underline{86.9}}$ & $85.9$ & $79.6$ &
  $92.0$ & $75.2$ & $75.4$ & $81.3$ & ${\underline{82.5}}$ & $73.4$ & $84.9$ & $94.8$ & $22.2$ & $66.7$ & $65.0$ &
  ${\underline{69.6}}$ \\
\bottomrule
\end{tabular}%
}
\caption{
Post-training success rates under tool environment shift:
Agents are post-trained on $\mathcal{D}^{\text{tr}}_{\text{all}}$ under each harness with tool schema fixed to \texttt{v1.0}, and evaluated on $\mathcal{D}_{\text{all}}^{\text{te}}$ under the same harness but with schema \texttt{v1.1} (mild shift) or \texttt{v2.0} (stronger shift).
Zero-shot results are included for reference, where the \textbf{Harness} column refers to the test-time harness.
}
\label{tab:exp1_toolshift_extended}
\end{table}

\clearpage

\begin{table}[!htbp]
  \centering
  \resizebox{0.82\linewidth}{!}{
  \begin{tabular}{c c c *{7}{c}}
  \toprule
  \multirow{2}{*}[-1ex]{\textbf{Model}} & \multirow{2}{*}[-1ex]{\textbf{Algorithm}} &
  \multirow{2}{*}[-1ex]{\textbf{Harness}}
  & \multicolumn{2}{c}{$\mathcal{D}^{\text{te}}_{\text{easy}}$}
  & \multicolumn{3}{c}{$\mathcal{D}^{\text{te}}_{\text{med}}$}
  & \multicolumn{1}{c}{$\mathcal{D}^{\text{te}}_{\text{hard}}$}
  & \multirow{2}{*}[-1ex]{All} \\
  \cmidrule(lr){4-5} \cmidrule(lr){6-8} \cmidrule(lr){9-9}
  & & & Pick & Look & Clean & Heat & Cool & Pick 2 & \\
  \midrule
  \multirow{6}{*}{Qwen2.5-3B-Instruct}
  & \multirow{3}{*}{GRPO}                       & \texttt{h-low} & $74.6_{\pm 1.7}$ & $54.8_{\pm 3.2}$ & $0.6_{\pm 1.0}$ & $0.0_{\pm 0.0}$ & $2.9_{\pm 1.3}$ & $1.6_{\pm 2.8}$ & ${\underline{23.1}}_{\pm 0.6}$ \\
  &                       & \texttt{h-mid} & $84.7_{\pm 1.7}$ & $76.3_{\pm 3.7}$ & $6.9_{\pm 0.0}$ & $12.8_{\pm 6.8}$ & $9.4_{\pm 5.0}$ & $13.0_{\pm 3.7}$ & ${\underline{33.7}}_{\pm 1.4}$ \\
  &                       & \texttt{h-high} & $88.7_{\pm 2.6}$ & $71.0_{\pm 8.5}$ & $74.7_{\pm 5.0}$ & $62.4_{\pm 5.9}$ & $66.7_{\pm 3.3}$ & $0.8_{\pm 1.4}$ & ${\underline{63.1}}_{\pm 1.7}$ \\
  \cmidrule(lr){2-10}
  & \multirow{3}{*}{GiGPO}
  & \texttt{h-low} & $89.8_{\pm 1.4}$ & $88.2_{\pm 3.0}$ & $2.9_{\pm 0.8}$ & $0.0_{\pm 0.0}$ & $9.4_{\pm 4.1}$ & $38.2_{\pm 6.1}$ & ${\underline{37.2}}_{\pm 1.0}$ \\
  &                       & \texttt{h-mid} & $92.1_{\pm 1.6}$ & $90.3_{\pm 5.3}$ & $19.5_{\pm 2.9}$ & $39.3_{\pm 6.7}$ & $36.2_{\pm 3.7}$ & $4.9_{\pm 2.0}$ & ${\underline{46.6}}_{\pm 2.4}$ \\
 &                       & \texttt{h-high} & $93.8_{\pm 2.1}$ & $51.6_{\pm 2.6}$ & $30.5_{\pm 2.9}$ & $43.6_{\pm 2.1}$ & $50.7_{\pm 1.0}$ & $41.5_{\pm 2.0}$ & ${\underline{53.4}}_{\pm 1.0}$ \\
  \midrule
  \multirow{6}{*}{Qwen2.5-7B-Instruct}
  & \multirow{3}{*}{GRPO}
  & \texttt{h-low} & $64.4_{\pm 1.7}$ & $38.7_{\pm 3.2}$ & $36.2_{\pm 3.4}$ & $37.6_{\pm 5.3}$ & $36.2_{\pm 1.3}$ & $12.2_{\pm 2.4}$ & ${\underline{39.2}}_{\pm 1.7}$ \\
   &                       & \texttt{h-mid} & $77.4_{\pm 3.5}$ & $81.7_{\pm 6.7}$ & $0.6_{\pm 1.0}$ & $44.4_{\pm 6.5}$ & $36.2_{\pm 4.5}$ & $28.5_{\pm 5.1}$ & ${\underline{42.7}}_{\pm 1.0}$ \\
  &                       & \texttt{h-high} & $88.7_{\pm 1.0}$ & $90.3_{\pm 0.0}$ & $71.3_{\pm 2.6}$ & $63.2_{\pm 7.4}$ & $58.0_{\pm 3.3}$ & $56.1_{\pm 2.4}$ & ${\underline{71.5}}_{\pm 1.6}$ \\
  \cmidrule(lr){2-10}
  & \multirow{3}{*}{GiGPO}
  & \texttt{h-low} & $90.4_{\pm 1.0}$ & $86.0_{\pm 1.9}$ & $51.7_{\pm 0.0}$ & $34.2_{\pm 1.5}$ & $29.0_{\pm 5.5}$ & $30.1_{\pm 9.9}$ & ${\underline{54.4}}_{\pm 2.4}$ \\
  &                       & \texttt{h-mid} & $89.8_{\pm 1.7}$ & $90.3_{\pm 3.2}$ & $33.9_{\pm 1.0}$ & $35.0_{\pm 1.5}$   & $26.1_{\pm 4.3}$ & $61.0_{\pm 2.4}$ & ${\underline{55.2}}_{\pm 0.8}$ \\
  &                       & \texttt{h-high} & $88.1_{\pm 2.9}$ & $91.4_{\pm 1.9}$ & $63.2_{\pm 2.0}$ & $53.0_{\pm 5.3}$ & $65.2_{\pm 4.3}$ & $38.2_{\pm 3.7}$ & ${\underline{66.9}}_{\pm 0.8}$ \\
  \bottomrule
  \end{tabular}}
  \caption{
        Post-training success rates under task shift: Agents are post-trained on $\mathcal{D}^{\text{tr}}_{\text{easy}}$ under each harness with tool schema fixed to \texttt{v1.0}, and evaluated on $\mathcal{D}^{\text{te}}_{\text{all}}$ under the same harness and schema.
    }
  \label{tab:iid_posttraining_easy}
\end{table}

\begin{table}[!htbp]
  \centering
  \resizebox{0.82\linewidth}{!}{
  \begin{tabular}{c c c *{7}{c}}
  \toprule
  \multirow{2}{*}[-1ex]{\textbf{Model}} & \multirow{2}{*}[-1ex]{\textbf{Algorithm}} &
  \multirow{2}{*}[-1ex]{\textbf{Harness}}
  & \multicolumn{2}{c}{$\mathcal{D}^{\text{te}}_{\text{easy}}$}
  & \multicolumn{3}{c}{$\mathcal{D}^{\text{te}}_{\text{med}}$}
  & \multicolumn{1}{c}{$\mathcal{D}^{\text{te}}_{\text{hard}}$}
  & \multirow{2}{*}[-1ex]{All} \\
  \cmidrule(lr){4-5} \cmidrule(lr){6-8} \cmidrule(lr){9-9}
  & & & Pick & Look & Clean & Heat & Cool & Pick 2 & \\
  \midrule
  \multirow{6}{*}{Qwen2.5-3B-Instruct}
  & \multirow{3}{*}{GRPO}
  & \texttt{h-low} & $6.2_{\pm 2.6}$ & $0.0_{\pm 0.0}$ & $0.0_{\pm 0.0}$ & $0.0_{\pm 0.0}$ & $0.0_{\pm 0.0}$ & $0.0_{\pm 0.0}$ & ${\underline{1.3}}_{\pm 0.6}$ \\
  &         &              \texttt{h-mid} & $59.9_{\pm 1.0}$ & $2.2_{\pm 1.9}$ & $64.4_{\pm 5.0}$ & $57.3_{\pm 1.5}$ & $36.2_{\pm 6.6}$ & $3.3_{\pm 3.7}$ & ${\underline{41.5}}_{\pm 2.4}$ \\
  &                       & \texttt{h-high} & $81.9_{\pm 1.0}$ & $32.3_{\pm 3.2}$ & $81.0_{\pm 1.7}$ & $89.7_{\pm 0.0}$ & $85.5_{\pm 1.3}$ & $27.6_{\pm 7.8}$ & ${\underline{69.7}}_{\pm 1.1}$ \\
  \cmidrule(lr){2-10}
  & \multirow{3}{*}{GiGPO}
  & \texttt{h-low} & $6.2_{\pm 2.6}$ & $0.0_{\pm 0.0}$ & $0.0_{\pm 0.0}$ & $0.0_{\pm 0.0}$ & $0.0_{\pm 0.0}$ & $0.0_{\pm 0.0}$ & ${\underline{1.3}}_{\pm 0.6}$ \\
  &                       & \texttt{h-mid} & $53.1_{\pm 1.0}$ & $18.3_{\pm 4.9}$ & $81.0_{\pm 1.7}$ & $76.9_{\pm 2.6}$ & $81.2_{\pm 1.3}$ & $17.9_{\pm 3.7}$ & ${\underline{57.9}}_{\pm 1.1}$ \\
  &                        & \texttt{h-high} & $51.4_{\pm 3.9}$ & $12.9_{\pm 3.2}$ & $91.4_{\pm 1.7}$ & $86.3_{\pm 3.0}$ & $86.2_{\pm 2.5}$ & $10.6_{\pm 1.4}$ & ${\underline{60.2}}_{\pm 0.6}$ \\
  \midrule
  \multirow{6}{*}{Qwen2.5-7B-Instruct}
  & \multirow{3}{*}{GRPO}
  & \texttt{h-low} & $61.0_{\pm 1.7}$ & $11.8_{\pm 1.9}$ & $75.3_{\pm 3.6}$ & $73.5_{\pm 3.9}$ & $72.5_{\pm 2.5}$ & $35.0_{\pm 5.1}$ & ${\underline{58.3}}_{\pm 2.0}$ \\
    &                       & \texttt{h-mid} & $59.3_{\pm 4.5}$ & $28.0_{\pm 6.7}$ & $64.9_{\pm 2.0}$ & $70.1_{\pm 1.5}$ & $58.0_{\pm 2.5}$ & $26.0_{\pm 2.8}$ & ${\underline{53.3}}_{\pm 1.9}$ \\
  &                       & \texttt{h-high} & $86.4_{\pm 0.0}$ & $88.2_{\pm 3.7}$ & $89.1_{\pm 1.0}$ & $78.6_{\pm 3.9}$ & $79.7_{\pm 1.3}$ & $34.1_{\pm 4.2}$ & ${\underline{77.1}}_{\pm 0.4}$ \\
  \cmidrule(lr){2-10}
  & \multirow{3}{*}{GiGPO}
  & \texttt{h-low} & $54.8_{\pm 2.6}$ & $32.3_{\pm 3.2}$ & $73.0_{\pm 1.0}$ & $75.2_{\pm 3.0}$ & $79.7_{\pm 1.3}$ & $4.9_{\pm 2.4}$ & ${\underline{55.7}}_{\pm 0.2}$ \\
  &                       & \texttt{h-mid} & $68.4_{\pm 3.5}$ & $47.3_{\pm 8.1}$ & $71.8_{\pm 1.0}$ & $76.1_{\pm 3.9}$ & $67.4_{\pm 0.0}$ & $17.1_{\pm 2.4}$ & ${\underline{60.0}}_{\pm 0.6}$ \\
  &                       & \texttt{h-high} & $81.4_{\pm 3.4}$ & $23.7_{\pm 1.9}$ & $96.0_{\pm 1.0}$ & $82.1_{\pm 4.4}$ & $92.0_{\pm 3.3}$ & $52.0_{\pm 6.1}$ & ${\underline{75.4}}_{\pm 1.3}$ \\
  \bottomrule
  \end{tabular}}
  \caption{
    Post-training success rates under task shift: Agents are post-trained on $\mathcal{D}^{\text{tr}}_{\text{med}}$ under each harness with tool schema fixed to \texttt{v1.0}, and evaluated on $\mathcal{D}^{\text{te}}_{\text{all}}$ under the same harness and schema.
  For \texttt{Qwen2.5-3B-Instruct} under \texttt{h-low}, post-training fails to take effect, so zero-shot results are reported instead.
  }
  \label{tab:iid_posttraining_medium}
\end{table}

\begin{table}[!htbp]
  \centering
  \resizebox{0.82\linewidth}{!}{
  \begin{tabular}{c c c *{7}{c}}
  \toprule
  \multirow{2}{*}[-1ex]{\textbf{Model}} & \multirow{2}{*}[-1ex]{\textbf{Algorithm}} &
  \multirow{2}{*}[-1ex]{\textbf{Harness}}
  & \multicolumn{2}{c}{$\mathcal{D}^{\text{te}}_{\text{easy}}$}
  & \multicolumn{3}{c}{$\mathcal{D}^{\text{te}}_{\text{med}}$}
  & \multicolumn{1}{c}{$\mathcal{D}^{\text{te}}_{\text{hard}}$}
  & \multirow{2}{*}[-1ex]{All} \\
  \cmidrule(lr){4-5} \cmidrule(lr){6-8} \cmidrule(lr){9-9}
  & & & Pick & Look & Clean & Heat & Cool & Pick 2 & \\
  \midrule
  \multirow{6}{*}{Qwen2.5-3B-Instruct}
  & \multirow{3}{*}{GRPO}
  & \texttt{h-low} & $6.2_{\pm 2.6}$ & $0.0_{\pm 0.0}$ & $0.0_{\pm 0.0}$ & $0.0_{\pm 0.0}$ & $0.0_{\pm 0.0}$ & $0.0_{\pm 0.0}$ & ${\underline{1.3}}_{\pm 0.6}$ \\
  &                       & \texttt{h-mid} & $11.9_{\pm 4.5}$ & $2.2_{\pm 1.9}$ & $0.0_{\pm 0.0}$ & $3.4_{\pm 3.0}$ & $0.0_{\pm 0.0}$ & $0.0_{\pm 0.0}$ & ${\underline{3.3}}_{\pm 1.1}$ \\
  &                       & \texttt{h-high} & $35.0_{\pm 2.6}$ & $15.1_{\pm 1.9}$ & $11.5_{\pm 1.0}$ & $12.8_{\pm 0.0}$ & $2.2_{\pm 2.2}$ & $0.0_{\pm 0.0}$ & ${\underline{13.9}}_{\pm 0.7}$ \\
  \cmidrule(lr){2-10}
  & \multirow{3}{*}{GiGPO}
  & \texttt{h-low} & $6.2_{\pm 2.6}$ & $0.0_{\pm 0.0}$ & $0.0_{\pm 0.0}$ & $0.0_{\pm 0.0}$ & $0.0_{\pm 0.0}$ & $0.0_{\pm 0.0}$ & ${\underline{1.3}}_{\pm 0.6}$ \\
  &                       &  \texttt{h-mid} & $11.9_{\pm 4.5}$ & $2.2_{\pm 1.9}$ & $0.0_{\pm 0.0}$ & $3.4_{\pm 3.0}$ & $0.0_{\pm 0.0}$ & $0.0_{\pm 0.0}$ & ${\underline{3.3}}_{\pm 1.1}$ \\
  &                       & \texttt{h-high} & $35.0_{\pm 2.6}$ & $15.1_{\pm 1.9}$ & $11.5_{\pm 1.0}$ & $12.8_{\pm 0.0}$ & $2.2_{\pm 2.2}$ & $0.0_{\pm 0.0}$ & ${\underline{13.9}}_{\pm 0.7}$ \\
  \midrule
  \multirow{6}{*}{Qwen2.5-7B-Instruct}
  & \multirow{3}{*}{GRPO}
  & \texttt{h-low} & $80.8_{\pm 1.0}$ & $15.1_{\pm 9.9}$ & $13.8_{\pm 4.6}$ & $0.9_{\pm 1.5}$ & $4.3_{\pm 2.2}$ & $68.3_{\pm 4.2}$ & ${\underline{33.1}}_{\pm 0.2}$ \\
  &                       & \texttt{h-mid} & $71.8_{\pm 1.0}$ & $44.1_{\pm 1.9}$ & $13.8_{\pm 1.7}$ & $6.0_{\pm 3.9}$ & $13.0_{\pm 4.3}$ & $65.0_{\pm 2.8}$ & ${\underline{36.1}}_{\pm 0.0}$ \\
  &                       & \texttt{h-high} & $81.4_{\pm 1.7}$ & $53.8_{\pm 9.3}$ & $31.0_{\pm 3.0}$ & $29.1_{\pm1.5}$ & $24.6_{\pm 7.0}$ & $81.3_{\pm 1.4}$ & ${\underline{50.6}}_{\pm 1.2}$ \\
  \cmidrule(lr){2-10}
  & \multirow{3}{*}{GiGPO}
  & \texttt{h-low} & $41.2_{\pm 1.0}$ & $17.2_{\pm 4.9}$ & $15.5_{\pm 0.0}$ & $3.4_{\pm 1.5}$ & $0.0_{\pm 0.0}$ & $95.1_{\pm 0.0}$ & ${\underline{28.8}}_{\pm 0.4}$ \\
  &                       & \texttt{h-mid} & $81.4_{\pm 1.7}$ & $31.2_{\pm 1.9}$ & $7.5_{\pm 1.0}$ & $18.8_{\pm 3.9}$ & $5.8_{\pm 1.3}$ & $83.7_{\pm 3.7}$ & ${\underline{38.8}}_{\pm 0.6}$ \\
  &                       & \texttt{h-high} & $95.5_{\pm 1.0}$ & $93.5_{\pm 3.2}$ & $54.0_{\pm 7.0}$ & $56.4_{\pm 4.4}$ & $51.4_{\pm 5.5}$ & $94.3_{\pm 3.7}$ & ${\underline{73.4}}_{\pm 2.9}$ \\
  \bottomrule
  \end{tabular}}
  \caption{
    Post-training success rates under task shift: Agents are post-trained on $\mathcal{D}^{\text{tr}}_{\text{hard}}$ under each harness with tool schema fixed to \texttt{v1.0}, and evaluated on $\mathcal{D}^{\text{te}}_{\text{all}}$ under the same harness and schema.
    For \texttt{Qwen2.5-3B-Instruct}, post-training fails to take effect, so zero-shot results are reported instead.
  }
  \label{tab:iid_posttraining_hard}
\end{table}

\clearpage

\color{black}

\section{Ablation Study}

\subsection{Generalization to a Different Agentic Environment}
\label{sec:webshop}

To examine whether our findings extend beyond \texttt{ALFWorld}, we additionally conduct experiments on \texttt{WebShop} \cite{yao2022webshop}, an interactive e-commerce environment in which an agent must purchase a product satisfying a user request. While \texttt{ALFWorld} focuses on household planning, \texttt{WebShop} requires stateful interaction with an e-commerce interface, providing a complementary setting for evaluating tool-integrated agents. Since \texttt{WebShop} does not provide task-type annotations analogous to the task categories in \texttt{ALFWorld}, we focus on tool environment shift.

\paragraph{Tool Schema.}
We construct two tool schemas while preserving the underlying task and available functionality. Under the base schema \texttt{v1.0}, the agent is provided with a \texttt{Search} tool for retrieving products and a general \texttt{Click} tool for interacting with clickable elements, including selecting product options, viewing product details, purchasing an item, and navigating between pages. Under the shifted schema \texttt{v2.0}, \texttt{Search} remains unchanged, while the general \texttt{Click} tool is decomposed into five specialized tools according to the semantic role of each clickable element. For example, selecting a product option requires \texttt{ChooseOption} rather than the generic \texttt{Click} tool. Thus, the task semantics and functionality remain fixed, while the action interface changes.

\paragraph{Harness.}
We instantiate two \texttt{WebShop}-specific harnesses, \texttt{h-low} and \texttt{h-high}. Under \texttt{h-low}, the agent must infer the currently selected product options from the interaction history. \texttt{h-high} additionally appends the currently selected options (\texttt{Selected:}) to every per-step observation, explicitly exposing the task-relevant interaction state. This plays a role analogous to the \texttt{Carrying:} information provided by \texttt{h-high} in \texttt{ALFWorld}. Tool descriptions are identical across the two harnesses. In addition, because the \texttt{WebShop} page itself exposes clickable elements, both harnesses provide the currently admissible actions: removing this information would introduce artificial partial observability rather than a realistic low-information harness.

\paragraph{Training and Evaluation Protocol.}
Following the tool environment shift protocol in \S\ref{sec:eval_protocol}, we post-train \texttt{Qwen2.5-3B-Instruct} with both GRPO and GiGPO under each harness, with the training tool schema fixed to \texttt{v1.0}. We then evaluate each agent under the same harness using either \texttt{v1.0} for in-distribution evaluation or \texttt{v2.0} for tool environment shift. We evaluate on a held-out split of 500 instructions over three seeds. We report the mean \texttt{WebShop} task score, success rate (SR), and the average number of tool-calling turns, with a maximum of 15 turns per episode.

\paragraph{Results.}
The \texttt{WebShop} results reproduce the main trend observed in \texttt{ALFWorld}: post-training with the more informative harness yields stronger performance under tool environment shift (\cref{tab:rebuttal_toolshift_score_turns}). Across both GRPO and GiGPO, agents post-trained under \texttt{h-high} achieve higher performance under \texttt{v2.0} than their \texttt{h-low} counterparts.
The effect is most pronounced with GiGPO. Under \texttt{h-low}, the success rate drops from $54.7$ to $33.5$, whereas under \texttt{h-high} it decreases only from $65.3$ to $61.9$. The average number of turns also increases substantially more under \texttt{h-low} ($5.0 \rightarrow 10.6$) than under \texttt{h-high} ($5.4 \rightarrow 7.4$). These results indicate that the benefit of harness-aware post-training under a tool environment shift extends beyond \texttt{ALFWorld} to another agentic environment.

\begin{table}[H]
\centering
\begin{tabular}{c c c c c c c c}
\toprule
\multirow{2}{*}{\textbf{Algorithm}}
& \multirow{2}{*}{\textbf{Harness}}
& \multicolumn{3}{c}{\textbf{v1.0}}
& \multicolumn{3}{c}{\textbf{v2.0}} \\
\cmidrule(lr){3-5} \cmidrule(lr){6-8}
&
& Score & SR & Turns
& Score & SR & Turns \\
\midrule
\multirow{2}{*}{GRPO}
& \texttt{h-low}
& $71.4_{\pm 0.5}$ & $57.5_{\pm 1.0}$ & $6.8_{\pm 0.0}$
& $54.2_{\pm 0.7}$ & $43.4_{\pm 1.6}$ & $10.0_{\pm 0.1}$ \\
& \texttt{h-high}
& $78.9_{\pm 0.2}$ & $63.7_{\pm 1.1}$ & $6.2_{\pm 0.1}$
& $57.9_{\pm 0.5}$ & $48.7_{\pm 0.2}$ & $9.6_{\pm 0.1}$ \\
\cmidrule(lr){1-8}
\multirow{2}{*}{GiGPO}
& \texttt{h-low}
& $75.6_{\pm 0.1}$ & $54.7_{\pm 0.3}$ & $5.0_{\pm 0.0}$
& $46.9_{\pm 0.6}$ & $33.5_{\pm 1.1}$ & $10.6_{\pm 0.1}$ \\
& \texttt{h-high}
& $82.1_{\pm 0.5}$ & $65.3_{\pm 0.3}$ & $5.4_{\pm 0.1}$
& $74.3_{\pm 0.7}$ & $61.9_{\pm 1.6}$ & $7.4_{\pm 0.1}$ \\
\bottomrule
\end{tabular}%
\caption{
Post-training performance of $\texttt{Qwen2.5-3B-Instruct}$ on $\texttt{WebShop}$ under tool environment shift.
We report the score, success rate (SR), and the average number of turns across three random seeds.
}
\label{tab:rebuttal_toolshift_score_turns}
\end{table}

\clearpage

\subsection{Short Adaptation for Post-hoc Harness Application}
We further examine whether a short adaptation stage can close the performance gap of post-hoc harness application. Starting from \texttt{Qwen2.5-3B-Instruct} trained under \texttt{h-low}, we continue post-training under the target harness for 100 additional steps, 20\% of the main training budget (\cref{tab:posthoc_adaptation}).
Across both GRPO and GiGPO, short adaptation improves over naive post-hoc application, but training with the target harness from the beginning remains the strongest. For example, with GiGPO under \texttt{h-high}, adaptation improves success rate from $67.6$ to $75.9$, while training-time application achieves $82.1$. A similar trend holds for GRPO ($59.6 \rightarrow 63.3$ vs.\ $69.7$). 
\begin{table}[!htbp]
  \centering
  \begin{tabular}{c c c c}
  \toprule
  \textbf{Algorithm}
  & \textbf{Harness Application}
  & \texttt{h-mid}
  & \texttt{h-high} \\
  \midrule

  \multirow{3}{*}{GRPO}
  & Post-hoc (naive)
  & $57.5_{\pm 1.2}$
  & $59.6_{\pm 0.9}$ \\

  & Post-hoc (adapted)
  & $58.2_{\pm 1.3}$
  & $63.3_{\pm 0.2}$ \\

  & Training-time
  & $\mathbf{61.7}_{\pm 1.0}$
  & $\mathbf{69.7}_{\pm 0.6}$ \\

  \cmidrule(lr){1-4}

  \multirow{3}{*}{GiGPO}
  & Post-hoc (naive)
  & $69.3_{\pm 0.0}$
  & $67.6_{\pm 1.1}$ \\

  & Post-hoc (adapted)
  & $71.5_{\pm 0.6}$
  & $75.9_{\pm 0.4}$ \\

  & Training-time
  & $\mathbf{83.2}_{\pm 1.3}$
  & $\mathbf{82.1}_{\pm 2.4}$ \\

  \bottomrule
  \end{tabular}
  \caption{
  Naive post-hoc, short-adaptation, and training-time harness application:
  in-distribution success rates on $\mathcal{D}^{\text{te}}_{\text{all}}$
  for \texttt{Qwen2.5-3B-Instruct} post-trained with GRPO or GiGPO.
  For each target harness $\{\texttt{h-mid}, \texttt{h-high}\}$,
  \textit{Post-hoc (naive)} directly evaluates the agent post-trained under
  \texttt{h-low} using the target harness;
  \textit{Post-hoc (adapted)} further adapts the \texttt{h-low}-trained
  checkpoint to the target harness for 100 steps (20\% of the main training steps),
  selecting the best checkpoint on the validation set;
  and \textit{Training-time} denotes agents post-trained under the target harness
  from the beginning.
  }
  \label{tab:posthoc_adaptation}
\end{table}


\subsection{Fine-grained Task Shift Analysis}
We further evaluate task shift at the single-category level using \texttt{Qwen2.5-3B-Instruct}, by post-training agents on \texttt{Pick} or \texttt{Clean} only and evaluating on all task categories (\cref{tab:single_category_taskshift}). Across both GRPO and GiGPO, more informative harnesses tend to improve transfer to unseen categories. For example, with GiGPO trained on \texttt{Pick}, \texttt{h-high} improves overall success from $47.4$ to $68.5$ relative to \texttt{h-low}, including a
substantial gain on held-out \texttt{Pick 2} ($6.5 \rightarrow 52.8$). Although some individual categories exhibit non-monotonic behavior, the aggregate trend aligns with Observation~\ref{obs:ts}: informative harnesses support inter-task transfer even from a single training category.
\begin{table}[htbp!]
  \centering
  \resizebox{0.82\linewidth}{!}{
  \begin{tabular}{c c c *{7}{c}}
  \toprule
  \multirow{2}{*}[-1ex]{\textbf{Training Split}}
  & \multirow{2}{*}[-1ex]{\textbf{Algorithm}}
  & \multirow{2}{*}[-1ex]{\textbf{Harness}}
  & \multicolumn{2}{c}{$\mathcal{D}^{\text{te}}_{\text{easy}}$}
  & \multicolumn{3}{c}{$\mathcal{D}^{\text{te}}_{\text{med}}$}
  & \multicolumn{1}{c}{$\mathcal{D}^{\text{te}}_{\text{hard}}$}
  & \multirow{2}{*}[-1ex]{All} \\
  \cmidrule(lr){4-5} \cmidrule(lr){6-8} \cmidrule(lr){9-9}
  & & & Pick & Look & Clean & Heat & Cool & Pick 2 & \\
  \midrule

  \multirow{6}{*}{$\mathcal{D}^{\text{tr}}_{\texttt{Pick}}$}

  & \multirow{3}{*}{GRPO}
  & \texttt{h-low}
  & $83.1_{\pm 1.7}$ & $12.9_{\pm 3.2}$
  & $3.4_{\pm 1.7}$ & $3.4_{\pm 1.5}$ & $10.1_{\pm 4.5}$
  & $26.8_{\pm 6.5}$
  & ${\underline{26.3}}_{\pm 0.4}$ \\

  & & \texttt{h-mid}
  & $83.1_{\pm 2.9}$ & $19.4_{\pm 6.5}$
  & $50.6_{\pm 1.0}$ & $30.8_{\pm 2.6}$ & $26.8_{\pm 3.3}$
  & $0.0_{\pm 0.0}$
  & ${\underline{39.7}}_{\pm 0.2}$ \\

  & & \texttt{h-high}
  & $76.8_{\pm 4.3}$ & $44.1_{\pm 1.9}$
  & $35.1_{\pm 2.6}$ & $24.8_{\pm 5.9}$ & $21.7_{\pm 2.2}$
  & $56.9_{\pm 2.8}$
  & ${\underline{44.6}}_{\pm 0.4}$ \\

  \cmidrule(lr){2-10}

  & \multirow{3}{*}{GiGPO}
  & \texttt{h-low}
  & $93.8_{\pm 2.1}$ & $43.0_{\pm 3.0}$
  & $51.1_{\pm 1.6}$ & $27.4_{\pm 8.7}$ & $39.9_{\pm 3.7}$
  & $6.5_{\pm 2.3}$
  & ${\underline{47.4}}_{\pm 0.8}$ \\

  & & \texttt{h-mid}
  & $94.4_{\pm 2.9}$ & $17.2_{\pm 3.0}$
  & $41.4_{\pm 6.5}$ & $46.2_{\pm 3.6}$ & $31.9_{\pm 1.0}$
  & $27.6_{\pm 1.1}$
  & ${\underline{47.1}}_{\pm 2.1}$ \\

  & & \texttt{h-high}
  & $91.0_{\pm 2.1}$ & $52.7_{\pm 1.5}$
  & $80.5_{\pm 3.3}$ & $53.8_{\pm 6.3}$ & $61.6_{\pm 2.0}$
  & $52.8_{\pm 3.0}$
  & ${\underline{68.5}}_{\pm 1.5}$ \\

  \midrule

  \multirow{6}{*}{$\mathcal{D}^{\text{tr}}_{\texttt{Clean}}$}

  & \multirow{3}{*}{GRPO}
  & \texttt{h-low}
  & $6.2_{\pm 2.6}$ & $0.0_{\pm 0.0}$
  & $0.0_{\pm 0.0}$ & $0.0_{\pm 0.0}$ & $0.0_{\pm 0.0}$
  & $0.0_{\pm 0.0}$
  & ${\underline{1.3}}_{\pm 0.6}$ \\

  & & \texttt{h-mid}
  & $61.6_{\pm 2.0}$ & $28.0_{\pm 1.9}$
  & $63.2_{\pm 1.0}$ & $32.5_{\pm 5.3}$ & $41.3_{\pm 3.8}$
  & $2.4_{\pm 2.4}$
  & ${\underline{41.7}}_{\pm 0.8}$ \\

  & & \texttt{h-high}
  & $83.6_{\pm 1.0}$ & $28.0_{\pm 3.7}$
  & $89.1_{\pm 3.6}$ & $59.8_{\pm 5.3}$ & $64.5_{\pm 5.5}$
  & $0.8_{\pm 1.4}$
  & ${\underline{59.5}}_{\pm 1.7}$ \\

  \cmidrule(lr){2-10}

  & \multirow{3}{*}{GiGPO}
  & \texttt{h-low}
  & $6.2_{\pm 2.6}$ & $0.0_{\pm 0.0}$
  & $0.0_{\pm 0.0}$ & $0.0_{\pm 0.0}$ & $0.0_{\pm 0.0}$
  & $0.0_{\pm 0.0}$
  & ${\underline{1.3}}_{\pm 0.6}$ \\

  & & \texttt{h-mid}
  & $71.8_{\pm 2.6}$ & $5.4_{\pm 1.9}$
  & $91.4_{\pm 1.7}$ & $74.4_{\pm 6.8}$ & $77.5_{\pm 4.5}$
  & $18.7_{\pm 6.1}$
  & ${\underline{61.8}}_{\pm 1.1}$ \\

  & & \texttt{h-high}
  & $73.4_{\pm 2.1}$ & $41.9_{\pm 4.6}$
  & $90.8_{\pm 1.6}$ & $52.1_{\pm 3.2}$ & $74.6_{\pm 1.0}$
  & $36.6_{\pm 2.0}$
  & ${\underline{65.2}}_{\pm 2.0}$ \\

  \bottomrule
  \end{tabular}
  }
  \caption{
  Post-training success rates under task shift at the single-category level.
  \texttt{Qwen2.5-3B-Instruct} agents are post-trained with GRPO or GiGPO
  on training instances from a single task category
  ($\texttt{Pick}$ or $\texttt{Clean}$) under each harness,
  with tool schema fixed to \texttt{v1.0}, and evaluated on
  $\mathcal{D}^{\text{te}}_{\text{all}}$ under the same harness and schema.
  For the \texttt{Clean}-trained agents under \texttt{h-low},
  post-training fails to take effect, so zero-shot results are reported instead.
  }
  \label{tab:single_category_taskshift}
\end{table}

\end{document}